\documentclass{article}

\usepackage{arxiv}

\usepackage[utf8]{inputenc} 
\usepackage[T1]{fontenc}    
\usepackage{hyperref}       
\usepackage{url}            
\usepackage{booktabs}       
\usepackage{amsfonts}       
\usepackage{nicefrac}       
\usepackage{microtype}      
\usepackage{lipsum}
\usepackage{graphicx}
\graphicspath{ {./images/} }

\usepackage{amsmath} 
\usepackage{amssymb} 
\usepackage{graphicx}
\usepackage{float} 

\usepackage{amsfonts}
\usepackage{algorithm}
\usepackage{algpseudocode}
\usepackage{titlesec} 
\usepackage{multirow} 
\usepackage{booktabs} 
\usepackage{lipsum} 
\usepackage{subcaption}

\title{Hypergraph-based multi-scale spatio-temporal graph convolution network for Time-Series anomaly detection}

\author{
 Hongyi Xu \\
  Edinburgh Parallel Computing Center\\
  University of Edinburgh\\
  47 Potterrow, Edinburgh\\
  EH8 9BT UK\\
  \texttt{s2044819@ed.ac.uk} \\
}

\begin{document}
\maketitle
\begin{abstract}
Multivariate time series anomaly detection technology plays an important role in many fields including aerospace ~\cite{7486356}, water treatment ~\cite{8215783}, cloud service providers ~\cite{9313014}, etc. Excellent anomaly detection models can greatly improve work efficiency and avoid major economic losses. However, with the development of technology, the increasing size and complexity of data, and the lack of labels for relevant abnormal data, it is becoming increasingly challenging to perform effective and accurate anomaly detection in high-dimensional and complex data sets. In this paper, we propose a hypergraph-based spatiotemporal graph convolutional neural network model STGCN\_Hyper, which explicitly captures high-order, multi-hop correlations between multiple variables through a hypergraph based dynamic graph structure learning module. On this basis, we further use the hypergraph based spatiotemporal graph convolutional network to utilize the learned hypergraph structure to effectively propagate and aggregate one-hop and multi-hop related node information in the convolutional network, thereby obtaining rich spatial information. Furthermore, through the multi-scale TCN dilated convolution module, the STGCN\_hyper model can also capture the dependencies of features at different scales in the temporal dimension. An unsupervised anomaly detector based on PCA and GMM is also integrated into the STGCN\_hyper model. Through the anomaly score of the detector, the model can detect the anomalies in an unsupervised way. Experimental results on multiple time series datasets show that our model can flexibly learn the multi-scale time series features in the data and the dependencies between features, and outperforms most existing baseline models in terms of precision, recall, F1-score on anomaly detection tasks. Our code is available on: \url{https://git.ecdf.ed.ac.uk/msc-23-24/s2044819}
\end{abstract}



\section{Introduction}
\textbf{Motivations.} In today's digital information age, anomaly detection and diagnosis in complex systems have become critical tasks to ensure the safety of corresponding infrastructures, which is crucial for improving work efficiency and avoiding significant economic losses. Anomaly detection tasks based on time series data are widely applied across various practical fields, such as monitoring the status of industrial equipment~\cite{6882174}, detecting financial fraud in banking~\cite{HILAL2022116429}, and effectively monitoring and providing early warnings for servers in large-scale cloud operators~\cite{9313014}. However, with the rapid development of infrastructure in various industries (for example, the development of sensors for monitoring specialized equipment in the industrial field), in most cases, industrial systems generate and transmit large volumes of time series data from multiple, even thousands of, sensors simultaneously at a single point in time, forming complex multivariate time series data (e.g. cloud server clusters, sensors in water treatment plants). Therefore, manually recording data and relying on human labor for anomaly detection is impractical. Thus, the implementation of an anomaly detection model capable of detecting, forecasting, and diagnosing anomaly signals in large-scale systems has become an important research area, as it can help prevent significant economic losses and ensure the safety of industrial equipment and associated personnel.

\textbf{Existing Solutions and Challenges.} However, with the advancement of contemporary science and technology and the proliferation of the Internet of Things (IoT), the types of data have become increasingly complex, and the scale of data has grown exponentially. As a result, effective anomaly detection in large-scale industrial equipment or infrastructure clusters has become increasingly difficult. Additionally, for multivariate time series datasets, the lack of labeled data (due to high manual costs) and the diversity and real-time dynamic changes of anomalies pose even greater challenges. Traditional supervised learning approaches have become less feasible in this domain. \textbf{(Challenge I)} 

Although with the rapid development of technology, many unsupervised learning-based anomaly detection models have emerged on the market, such as traditional statistical unsupervised learning models: 
\begin{enumerate}
    \item Clustering-based models like SSC-OCSVM~\cite{9147152}, which learn the normal distribution of normal time series window data and treat data points that deviate from the normal distribution as anomalies.
    \item Distance-based models like Matrix Profile I and density estimation-based models (LOF and LFCOF).
\end{enumerate}

These approaches fail to effectively capture the temporal correlations between variables and the interdependencies among high-dimensional variables. 

Additionally, some deep learning-based models, such as neural networks based on LSTM \cite{8836638} or RNN \cite{7486356}, can enhance the model's ability to detect anomalies to some extent. However, these current neural networks based on time series structures also have corresponding issues:

\begin{enumerate}
    \item They only consider the temporal dependencies of variables without considering the interdependencies between variables or the higher-order relationships among variables \textbf{(Challenge II)}.
    
    \item Sequence models based on LSTM and RNN become less useful when dealing with large amounts of data, as these recursive models cannot efficiently parallelize computations and require processing all preceding units' information before predicting the next time step \textbf{(Challenge III)}.
    
    \item Moreover, although existing graph-based models on the market can be used to model pairwise dependencies among high-dimensional time series variables, traditional Graph Convolutional Networks (GCNs) \cite{9020760} only adopt pairwise (binary) connections between variables. However, the data correlations in real-world multimodal data can be much more complex; the relationships between variables may be ternary, quaternary, or even higher-order. Therefore, conventional GCN models have significant limitations in modeling such complex relationships \textbf{(Challenge IV)}. 
    
    \item Finally, applying Graph Neural Networks (GNNs) or Spatio-Temporal Neural Networks for anomaly detection on multivariate time series datasets and modeling the correlations between variables requires a predefined graph structure. This is necessary for nodes to transmit and aggregate information according to the learned topological structure between nodes, thereby enabling each variable to learn a good feature representation \textbf{(Challenge V)}.  
\end{enumerate}

\textbf{Contributions.} 
To address these challenges, we propose a Spatio-Temporal Graph Convolutional Neural Network based on a hypergraph structure (STGCN\_hyper) in this project. According to experimental results on multiple multivariate time series datasets, our model's anomaly detection capabilities, measured by Precision, Recall, and the harmonic mean F1-score, significantly outperform existing baseline models.

The main contributions of this paper are as follows:
\begin{itemize}

    \item Specifically, to tackle \textbf{Challenge I}, we introduce a prediction-based model that forecasts the next (or multiple) time steps' values using preceding time steps' values of each variable (as well as those of related variables). We establish an anomaly detector based on PCA or GMM through reconstruction errors. This detector models the probability distribution of normal data by analyzing the distribution of training and validation data in a low-dimensional space. Subsequently, it computes the likelihood of test data corresponding to the normal distribution in the low-dimensional space for anomaly detection, thereby avoiding reliance on labeled training data.
    
    \item To overcome \textbf{Challenge II} and \textbf{Challenge IV}, our proposed Spatio-Temporal Graph Neural Network model is built upon a hypergraph structure. Unlike conventional graph structures—where each edge (representing a relationship) can only connect two feature vertices—a hypergraph's hyperedges can connect more than two feature vertices, meaning the degree of a hyperedge can exceed two. This flexible hyperedge structure facilitates more adaptable modeling of complex variable dependencies (which may involve multivariate associations and skip connections in multimodal time series datasets), enabling the capture of higher-order intricate correlations.
    
    \item Addressing \textbf{Challenge V}, we embed a dynamic learning module for graph structures (i.e., variable correlations) within the model. This time series variable correlation learning module progressively and dynamically learns the inter-variable relationships during model training, thereby enhancing information aggregation and propagation within the hypergraph structure.

    \item To resolve \textbf{Challenge III}, our proposed multi-scale feature extraction Temporal Convolutional Network (TCN) module enables parallel training through CUDA streams. It simultaneously extracts short-term, mid-term, and long-term feature patterns along the temporal dimension, enhancing anomaly detection capabilities while ensuring training efficiency.
    
\end{itemize}

\section{Related Work}

\subsection{Machine Learning-based Solutions}
In the field of machine learning-based anomaly detection, solutions are typically divided into three approaches based on input type: supervised, unsupervised, and semi-supervised anomaly detection. The essence of supervised anomaly detection is a classification problem, aiming to distinguish between anomalies and normal instances. On the other hand, unsupervised anomaly detection does not have access to exact labels for a given dataset. It achieves anomaly detection by identifying shared patterns among data instances and observing outliers. Additionally, semi-supervised anomaly detection accepts a dataset with either normal or anomalous labels and determines whether the target data for anomaly detection is normal or anomalous.

\begin{itemize}
        \item \textbf{Supervised}: Training techniques under the fully supervised model require a labeled training dataset where instances are labeled as either normal or anomalous. The model then establishes a predictive model for both normal and anomalous classes. Any unseen data instance is compared against the model to determine which class it belongs to.
        
        \item \textbf{Semi-supervised}: In this mode, the training dataset used only includes instances labeled as the normal class, without the need for labels from the anomalous class. Therefore, semi-supervised anomaly detection constitutes a one-class classification problem. Clustering is the most representative method in this category: the goal of clustering is to group similar data into multiple clusters based on the machine learning task~\cite{LI2021106919}. The normal class is characterized by several prototype points in the input dataset. The class label of a test point is quantified by the distance to the nearest cluster center. In anomaly detection, normal data with similar behavior will be located in the same cluster, while outliers will exhibit significant differences in this standard.
        
        \item \textbf{Unsupervised}: Unsupervised techniques are typically used when there is no prior knowledge of the dataset, so unsupervised methods do not require training data. Techniques in this category implicitly assume that normal instances are more frequent than anomalies in the test data. The anomaly detection method must analyze the dataset to infer the true concept of anomalies or make assumptions about that concept.
\end{itemize}

For the task of anomaly detection on multivariate time series datasets, supervised learning methods require large amounts of labeled data for training. Moreover, the models trained based on supervised learning only maintain the ability to distinguish between the types of anomalies defined in the dataset. Their effectiveness is very limited for time series datasets with highly dynamic data characteristics. Currently, most of the unsupervised or semi-supervised learning methods available on the market are more suitable for detection on time series datasets. Depending on the detection method used, related work can be divided into the following categories:

\begin{enumerate}
    \item \textbf{Forecasting-based model}: Some models use neural networks based on stacked LSTM structures to improve the accuracy of modeling. Among them, LSTM-NDT~\cite{10.1145/3292500.3330672} is primarily applied to datasets in the aerospace field, addressing issues where spacecraft system data is typically highly unstable, with monitored data often being heterogeneous, noisy, and high-dimensional. This model improves the accuracy and performance of anomaly detection in such datasets. The CNN-based DeepAnt~\cite{8581424}, on the other hand, captures data distribution along the time dimension more effectively by combining time series detectors with a CNN architecture, which allows it to better detect periodic and seasonal point anomalies in data. However, LSTM models based on deterministic modeling do not consider the randomness and uncertainty of the data, while the detection performance of the DeepAnt model on contextual and collective anomalies is inferior to its performance on point anomalies.

    \item \textbf{Reconstruction-based model}: DAGMM~\cite{zong2018deep} proposes a detection model based on a deep autoencoder, which utilizes the deep autoencoder to generate low-dimensional representations and reconstruction errors for each input data point. These are then further fed into a Gaussian Mixture Model (GMM) for anomaly detection, addressing, to some extent, the issue where data points in low-dimensional space fail to retain essential information. LSTM-VAE~\cite{8279425} employs a serially connected LSTM and VAE layer to project the multimodal observations at each time step and their temporal dependencies into the latent space, estimating the expected distribution of the multimodal input from the latent space representation and reconstructing it, thereby optimizing the modeling capability for high-dimensional data. However, the DAGMM model is designed primarily for general multivariate datasets, and when applied to time series datasets, it overlooks the temporal dependencies between data points. On the other hand, LSTM-VAE effectively captures the temporal dependencies between data points but neglects the spatial dependencies between different variables.

    \item \textbf{Graph-based model}: GDN~\cite{Zhang_Huang_Xu_Xia_Dai_Bo_Zhang_Zheng_2021} proposes a novel graph neural network based on an attention mechanism. By constructing a graph structure model, it can learn the dependency relationships between pairs of variables. Additionally, the model employs different parameters for different feature nodes within the GNN architecture, which helps improve detection performance, especially in scenarios like water treatment plant datasets that involve a large number of sensors with nonlinear dependencies between them. However, GDN is limited in that it can only capture the correlations between pairs of variables, and it is unable to further capture the relationships between multi-hop neighbors in the graph. Moreover, when applied to time series datasets, the model overlooks the temporal dependencies between variables, which further affects its detection performance on such datasets.

    \item \textbf{Representation Models}: DCdetector~\cite{10.1145/3580305.3599295} employs a deep CNN model with a dual attention mechanism, focusing on the representation of data along the time dimension. It uses contrastive learning to enhance the ability to distinguish between normal and abnormal signals. Positive and negative samples are derived from different perspectives of the same time series data, and the model differentiates anomalies by maximizing the representational differences between normal and abnormal samples. This type of model is particularly effective in handling signals that include noise, seasonality, and abrupt changes. However, similar to other models, DCdetector lacks the ability to capture potential spatial correlations between multiple variables, limiting its effectiveness in diagnosing anomalies and identifying the root cause.
\end{enumerate}

\subsection{Graph Learning}
A graph, also known as a network, is a data structure that can be used to model rich associations between entities. Graph is typically composed of vertices representing entities and edges representing relationships between entities. Graph Neural Networks (GNNs) are models based on the graph data structure for data analysis. Their core idea is to update the feature representation of target nodes by propagating information between neighboring nodes through message passing. For example, Graph Convolutional Networks (GCNs) can update their representations using predefined edge weights, while Graph Attention Networks (GATs)\cite{veličković2018graphattentionnetworks} can further automatically learn the weights of neighboring nodes to aggregate information and update the representation of the target node. Methods that model relationships between entities based on graph structures have been applied in many fields, such as:

\begin{enumerate}
    \item Modeling community structures to simulate the spread of information through graph structures~\cite{10.1145/3132847.3132925}.
    \item Applying subgraph classification to chemical tasks, such as compound classification~\cite{pham2017graphclassificationdeeplearning}.
    \item Graph recovery methods based on graph learning can be used to reconstruct the original graph structure in cases where some data is missing or incomplete~\cite{shrivastava2019gladlearningsparsegraph}.
\end{enumerate}

\subsection{Graph Structure Learning}
Using graph-based neural networks for modeling usually requires a predefined graph structure, which, along with the data, is fed into the neural network as input. This structure acts as the topological connection between nodes to facilitate message passing in the graph. However, a predefined graph structure is not always easily obtainable, which is also the case in our project. To address the challenge of the unavailability of predefined graph structures, recent methods have emerged that focus on graph structure learning.

The main idea of Graph Structure Learning (GSL) is to optimize the GNN and the graph structure simultaneously during the training process. For example, Sublime~\cite{10.1145/3485447.3512186} proposes a contrastive learning framework that generates a learning target from the raw data as an "anchor" and uses contrastive loss to maximize the consistency between the target graph and the anchor, thereby ultimately learning the relationships between data features.

However, the GSL method based on Sublime is suitable for datasets where relationships between nodes are nearly static, making its direct application to time series datasets potentially inappropriate.

\section{Problem Formulation}
Assume we have a multivariate time series data sequence \(X = \{x_1, x_2, \ldots, x_T\}\) consisting of \(T\) consecutive observations, where \(x_T \in \mathbb{R}^N\). Each feature vector \(x_T\) at time point \(t\) is composed of multiple dimensional feature values, i.e., \(x_T = \{x_1, x_2, \ldots, x_j\}\), where each \(x_j\) represents the value of the \(j\)-th feature dimension at time point \(T\). The fundamental tasks of the model are mainly divided into the following two parts:

\vspace{1em} 

\begin{itemize}
\item\textbf{Anomaly Detection}: For the input training time series sequence data \(T\), the number of time steps contained in this time series \(T\) is determined by the model's hyperparameters. Our goal is to use a sliding window for subsequent \(h\)-step multivariate prediction, where \(h \geq 1\). We predict the feature vector values \(x_{T+h}\) for the next \(h\) time steps based on the known fixed-size sliding window data \(T\).

\vspace{1em} 

Our model also needs to address the unsupervised real-time anomaly detection problem. The model must first learn the regular (general) distribution of values for different features from a large training set. Then, when receiving a prediction stream containing normal or abnormal data, the model should detect data points that deviate significantly from the learned normal data distribution in real-time, thereby achieving anomaly detection. The model makes decisions based only on past data and cannot reverse its previous decisions.

\vspace{1em} 

(The prediction needs to be \(\mathbf{Y} = \{y_T, \ldots, y_{T+h}\}\), where \(y_t \in \{0, 1\}\) to detect anomaly events within \(h\) time steps after \(T\). Similarly, based on \(\{x_{1+k}^{(i)}, x_{2+k}^{(i)}, \ldots, x_{T+k}^{(i)}\}\), \(k \in \mathbb{R}^+\), the next sequence \(x_{T+h+k}\) is predicted.

\vspace{1em} 

\item\textbf{Capturing Multivariate Dependencies:} Another objective of this project is to deeply capture the complex dependencies in multivariate time series data by learning the underlying Laplacian matrix and the hypergraph structure among features. Consider a set of $n$ time series variables $X = \{x_1, x_2, \dots, x_n\}$, where our goal is to construct a hypergraph structure $H \in \mathbb{R}^{n \times m}$, with $m$ being the number of hyperedges, and $H_{ij}$ representing the strength of the association between node $i$ and hyperedge $j$. Based on this hypergraph, we compute the normalized Laplacian matrix $L$. This representation allows the model to capture high-order, many-to-many complex relationships among the variables. We extract features through multi-level graph convolution operations. This approach enables the model to effectively aggregate multi-hop neighbor information, thereby learning richer spatiotemporal dependencies. Ultimately, the model utilizes the learned features for future state prediction and anomaly detection. Through this hypergraph-based and graph neural network approach, our model can more comprehensively understand the complex dependencies among variables within the system.

\end{itemize}

\section{Methodology}
\subsection{Data Preprocessing}

\subsubsection{Data Cleaning}
\begin{itemize}
\item \textbf{Step 1}: Handling Missing Values: For potential feature columns with entirely missing values, we fill them with 0. For columns with partially missing values, we use the mean to fill them. 

\item \textbf{Step 2}: For potential date and time formatted data (currently treated as string type data), I standardized the date format to "mm/dd/yyyy" and the time format to "'HH:MM:SS'", then merged them and created a datetime type data object. 

\item \textbf{Step 3}: For the column names of each attribute, we removed all redundant fields representing path names, making the column names more uniform and concise. 

\item \textbf{Step 4}: Overfitting is a common phenomenon where the trained model performs extremely well on the training samples but poorly on new, unseen samples; that is, the model's generalization ability is poor. To find the best model parameters that appropriately balance these two aspects, it is necessary to divide the data into a training set and a validation set. The training set is used to build models with multiple parameter settings, then each trained model is tested using the validation set through cross-validation. The optimal model parameter set is determined by the parameter set with the minimum validation error, and finally, the test set is used for the final generalization ability test of the trained model. In our model, the dataset is roughly divided according to a ratio of 70
\end{itemize}

\subsubsection{Normalization} Due to the different value ranges of different feature variables, retaining the original value ranges of different variables may result in inconsistent contributions of different features to the loss function, ultimately leading to the model performance being dominated by features with larger value ranges, thus affecting the final model performance. Therefore, the first step of our data preprocessing pipeline is data normalization. Data normalization involves scaling attribute values to have the same interval/scale, thus having the same importance. There are three normalization techniques: Z-score normalization~\cite{patro2015normalizationpreprocessingstage}, Min-Max normalization~\cite{patro2015normalizationpreprocessingstage}, and decimal scaling normalization. In this project, we used MinMax normalization for STGCN-based model.

The MinMax normalization process is as follows: Min-Max normalization (often called feature scaling) performs a linear transformation on the original data. For each feature, the minimum value of the feature is transformed to 0, the maximum value is transformed to 1, and every other value is transformed to a decimal between 0 and 1. The formula is as follows:
   
\vspace{-0.5em}   
\[
x_{\text{normalized}} = \frac{x - x_{\text{min}}}{x_{\text{max}} - x_{\text{min}}} \quad \text{\scriptsize{where} \ $x_{\text{min}}$: \text{Minimum}, \ $x_{\text{max}}$: \text{Maximum}}
\]
   
Min-Max normalization preserves the relationships between the original data values. The cost of having this bounded range is that the final standard deviation is smaller, which can suppress the impact of outliers. The comparison of the WADI dataset features before and after normalization is shown in Figure \ref{fig:wadi_normalized}:

\begin{figure}[H]
\centering
\includegraphics[width=\textwidth]{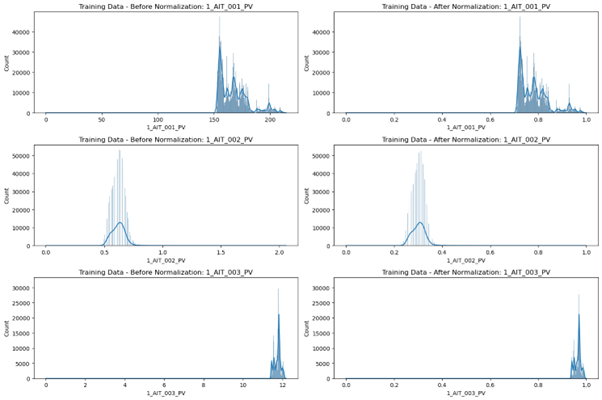} 
\caption{Normalization Effects}
\caption*{Note: This figure mainly shows the impact of Normalization on data distribution. We also take the three variables 1-AIT-001-PV, 1-AIT-002-PV, and 1-IT-003-PV as examples. The three figures on the left represent the data distribution before Normalization, and the three figures on the right represent the data distribution after Normalization.}
\label{fig:wadi_normalized}
\end{figure}

Figure \ref{fig:wadi_normalized} shows the comparison of three representative sensor data (\texttt{1\_AIT\allowbreak\_001\_PV}, \texttt{1\_AIT\allowbreak\_002\_PV}, and \texttt{1\_AIT\allowbreak\_003\_PV})  before and after normalization. For example, for the feature \texttt{1\_AIT\allowbreak\_001\_PV}, we observe that the original data is mainly distributed in the range of 0 to 200, showing a clear multimodal distribution. After normalization, the data is mapped to the range of 0 to 1 while retaining its original data distribution structure. This process not only unifies the data scale but also preserves the information that the feature may reflect about system states or operating modes. This normalization method lays the foundation for our subsequent anomaly detection model. It ensures the fairness of contributions from different features in the model while preserving the unique distribution characteristics of each feature.

\subsubsection{Sliding-windows Down-sampling} Given a time series \(T\) of length \(m\) and a user-defined subsequence length of \(n\), all possible subsequences can be extracted by sliding a window of size \(n\) across \(T\) and considering each subsequence \(C_p\) \cite{https://doi.org/10.1155/2014/879736}. In our model, we consider a fixed local context window of length \(K\), denoted as \(W_t = \{x_{t-K+1}, \ldots, x_t\}\). For \(t < K\), we use copy padding and convert the input time series \(T\) into sliding windows \(W = \{W_1, \ldots, W_t\}\). Through this preprocessing, we generate a series of input-output pairs, where the input data captures the short-term history of each variable, and the output data represents the future values we want the model to predict. This sliding window method not only enables the model to capture local patterns in the time series, but also provides structured input for graph structure learning and anomaly detection tasks.

We selected a window size of approximately 10 time steps, based on the characteristics of the WADI system and our focus on short- to mid-term patterns. The choice of window size balances capturing sufficient contextual information and maintaining computational efficiency. Larger windows can capture longer-term dependencies, while smaller windows focus more on local patterns. The sliding step is set to 1 time step, ensuring high overlap between windows to capture subtle changes in the data. Highly overlapping data windows help the model learn more continuous temporal dependencies, which is crucial for detecting gradual anomalies.

\subsection{The Architecture of ST-GCN based Model}
\subsubsection{Overall Architecture}
\begin{figure}[H]
\centering
\includegraphics[width=\textwidth]{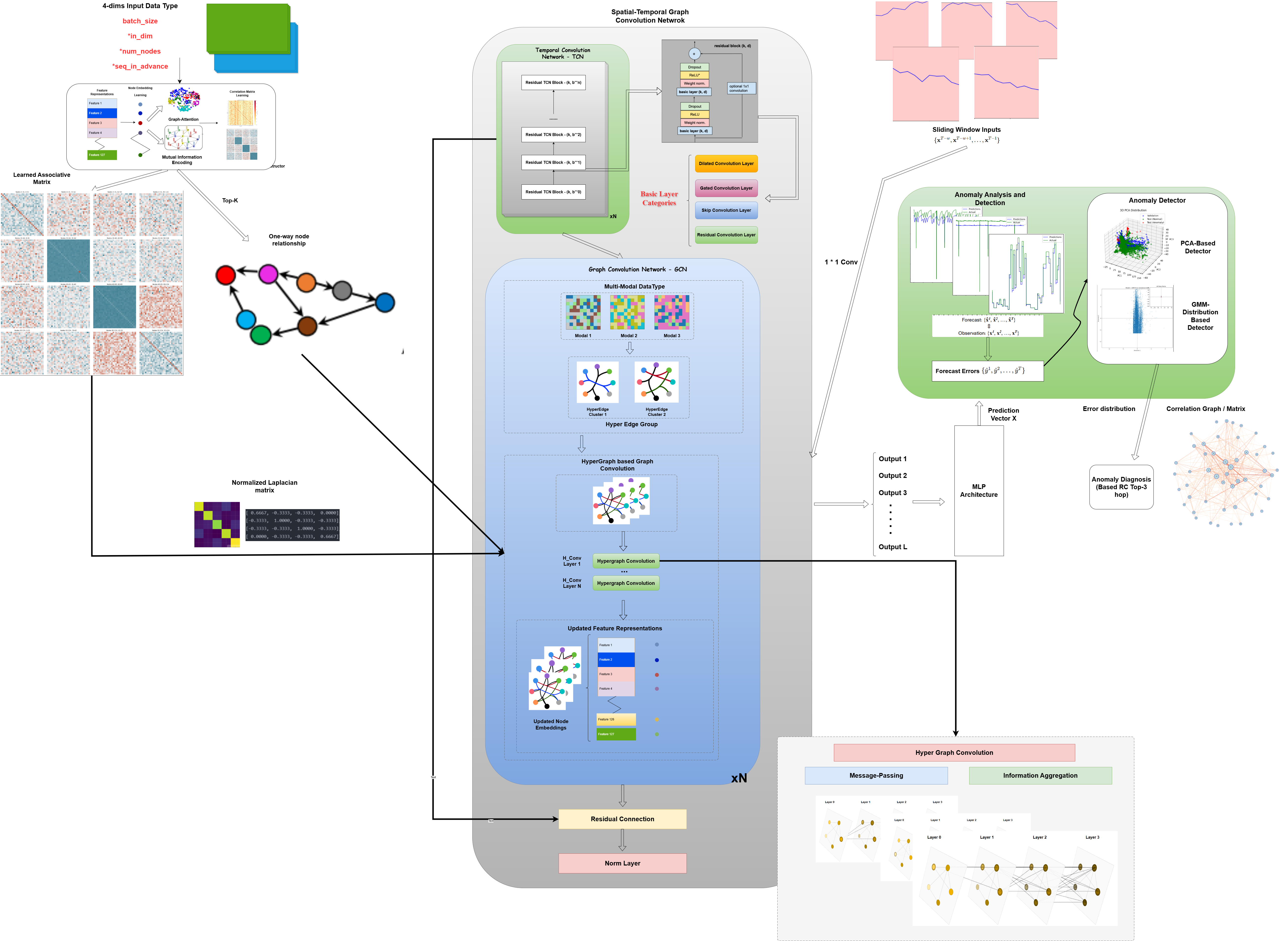} 
\caption{The Architecture of STGCN-Hyper Model}
\caption*{Note: \textbf{I}. Hypergraph Dynamic Structure Learning Module first initializes node embedding through the input data features, and then uses Mutual Infomation encoding to learn the normalized hypergraph Laplacian matrix based on the multivariate structure. The matrix will be passed into the STGCN graph convolution module for message passing operations. \textbf{II}. First, sliding windows of different scales are mapped to the target space through a 1*1 matrix, and then the features based on different time spans are captured through a parallel multi-core TCN convolution module, and then the correlation between multiple variables is captured through a hypergraph convolution module. Residual connections are used to prevent gradient disappearance and explosion. \textbf{III}. Output the predicted values of different features at the next time step (default) through a simple fully connected MLP structure layer. \textbf{IV}. By receiving the predicted value for the next time step, calculating the normalized prediction error with the true value, and mapping it to a low-dimensional space through PCA or GMM, the distribution of abnormal indicators is calculated through the log-likelihood, and abnormalities are reported for those data points that deviate significantly from the normal distribution.}
\label{fig:STGCN_architecture}
\end{figure}

In this section, we will detail the overall framework of STGCN-Hyper (Spatial-Temporal Graph Neural Network with Hypergraph) and its specific design for multivariate time series anomaly detection. As shown in Figure \ref{fig:STGCN_architecture}, our approach mainly consists of the following core components:

\begin{itemize}

\item I. Hypergraph Dynamic Structure Learning Module (MTCL), 
\item II. Hypergraph-based Spatial-Temporal Graph Convolutional Network, 
\item III. Anomaly Detection and Diagnosis Module. 

\end{itemize}

This model cleverly integrates Temporal Convolutional Networks (TCN), Graph Convolutional Networks (GCN), and dynamic hypergraph structures to capture temporal dynamics, spatial dependencies, and higher-order dependencies among multiple variables. The main advantage of the STGCN-Hyper model is its ability to adaptively learn potential multivariate dependencies and multi-hop relationships in the data, particularly excelling in handling nonlinear, non-stationary time series data.

\subsection{Key Components of the STGCN Model}
\subsubsection{Hypergraph-based Correlation Learning (MTCL)}

\hspace{2em}\textbf{Hypergraph Structure Learning}: To effectively model the higher-order dependencies between variables in complex systems, we designed an innovative multivariate mix-hop correlation learning layer based on hypergraph and hyperedge structures and proposed a method for adaptively learning the unknown hypergraph association matrix \(H\). In this framework, nodes represent variables, hyperedges represent higher-order connections between them. Unlike the concept of edges in ordinary graph structures, hyperedges allow for the connection of more than two vertices. Based on this property, we can capture dependencies among multiple variables while helping identify mix-hop neighbor nodes in the anomaly diagnosis module to find the root cause of anomalies. Specifically, our detailed formula is as follows:

\begin{equation}
\begin{cases}
    \mathbf{N}_1 = \tanh(\alpha \mathbf{N}_1 \mathbf{W}_1), \\
    \mathbf{N}_2 = \tanh(\alpha \mathbf{N}_2 \mathbf{W}_2), \\
    \mathbf{H} = \text{HyperedgeGenerator}(\mathbf{N}_1) \cdot \text{Attention}(\mathbf{N}_2),
\end{cases}
\end{equation}

where \(\mathbf{N}_1, \mathbf{N}_2 \in \mathbb{R}^{N \times d}\) represent the two initial random node~\cite{rozemberczki2021multi} embedding matrices, \(N\) denotes the number of variables, and \(d\) denotes the embedding dimension of each node. These matrices are continually adjusted during training to accurately reflect the representations of the features. \(\mathbf{W}_1, \mathbf{W}_2 \in \mathbb{R}^{d \times d}\) are the trainable matrices. \(\alpha\) represents the control of the degree of the nonlinear activation function, with larger \(\alpha\) values indicating a greater degree of nonlinearity. Subsequently, we apply the above calculation results to the activation function to capture nonlinear relationships:

\begin{equation}
\mathbf{H} = \text{ReLU}(\text{HyperedgeGenerator}(\mathbf{N}_1) \cdot \text{Attention}(\mathbf{N}_1))
\end{equation}

HyperedgeGenerator is a feedforward neural network that generates the weights of each hyperedge to describe the high-order relationship between nodes. The weight in the return parameter represents the connection strength between nodes.

\begin{equation}
\text{HyperedgeGenerator}(\mathbf{N}_1) = \mathbf{W}_h^{(2)} \cdot \text{ReLU}(\mathbf{W}_h^{(1)} \mathbf{N}_1)
\end{equation}

\begin{equation}
\text{Attention}(\mathbf{N}_1) = \sigma(\mathbf{W}_a \mathbf{N}_1)
\end{equation}

Subsequently, through the attention mechanism, the model can dynamically adjust the weight and importance of each node in different hyperedges. Then, through the Sigmoid activation function \(\sigma\), the model ultimately compresses the weight range to \([0, 1]\).

\vspace{1em} 

\hspace{1em}\textbf{Normalized Laplacian Matrix Construction}: Based on the \(H\) matrix constructed in Step 1, we first add an identity matrix \(I_N\) to reflect the inherent association of each node with itself. Then, based on the adjusted \(H\) matrix, we calculate the corresponding degree matrix:
\begin{equation}
\begin{cases}
    D_v = \text{diag}(\sum_j H_{ij}), \\
    D_e = \text{diag}(\sum_i H_{ij}),
\end{cases}
\end{equation}

where \(D_v\) is the node degree matrix, recording the number of hyperedges connected to each node, and \(D_e\) is the hyperedge degree matrix, recording the number of nodes connected by each hyperedge. 
Next, we perform the inverse square root of the degree matrices and finally compute the normalized Laplacian matrix:
\begin{equation}
\mathcal{L} = I_N - D_v^{-1/2} H D_e^{-1} H^T D_v^{-1/2}
\end{equation}

This process helps to balance the influence of different nodes and hyperedges during the calculation of the Laplacian matrix~\cite{merris1994laplacian}. The normalized Laplacian matrix can better reflect the relationships between nodes in the graph structure, which is beneficial for the feature extraction process in subsequent graph convolutions.

\begin{figure}[H]
    \centering
    \resizebox{0.99\textwidth}{!}{ 
    \begin{minipage}{0.3\textwidth}
        \centering
        \includegraphics[width=0.95\textwidth]{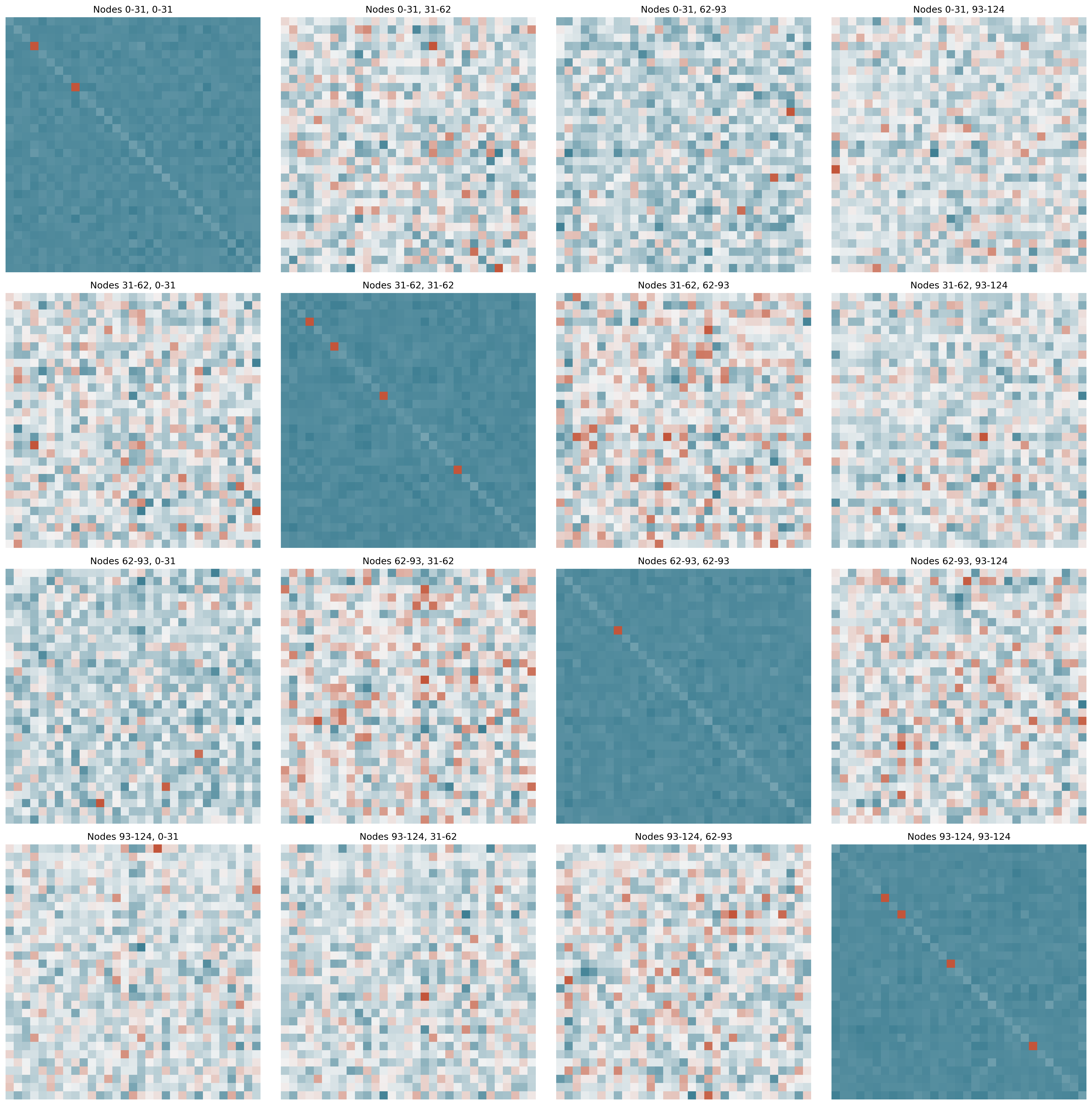} 
        \caption*{Epoch 1}
        \label{fig:Laplacian1}
    \end{minipage}%
    
    \begin{minipage}{0.3\textwidth}
        \centering
        \includegraphics[width=0.95\textwidth]{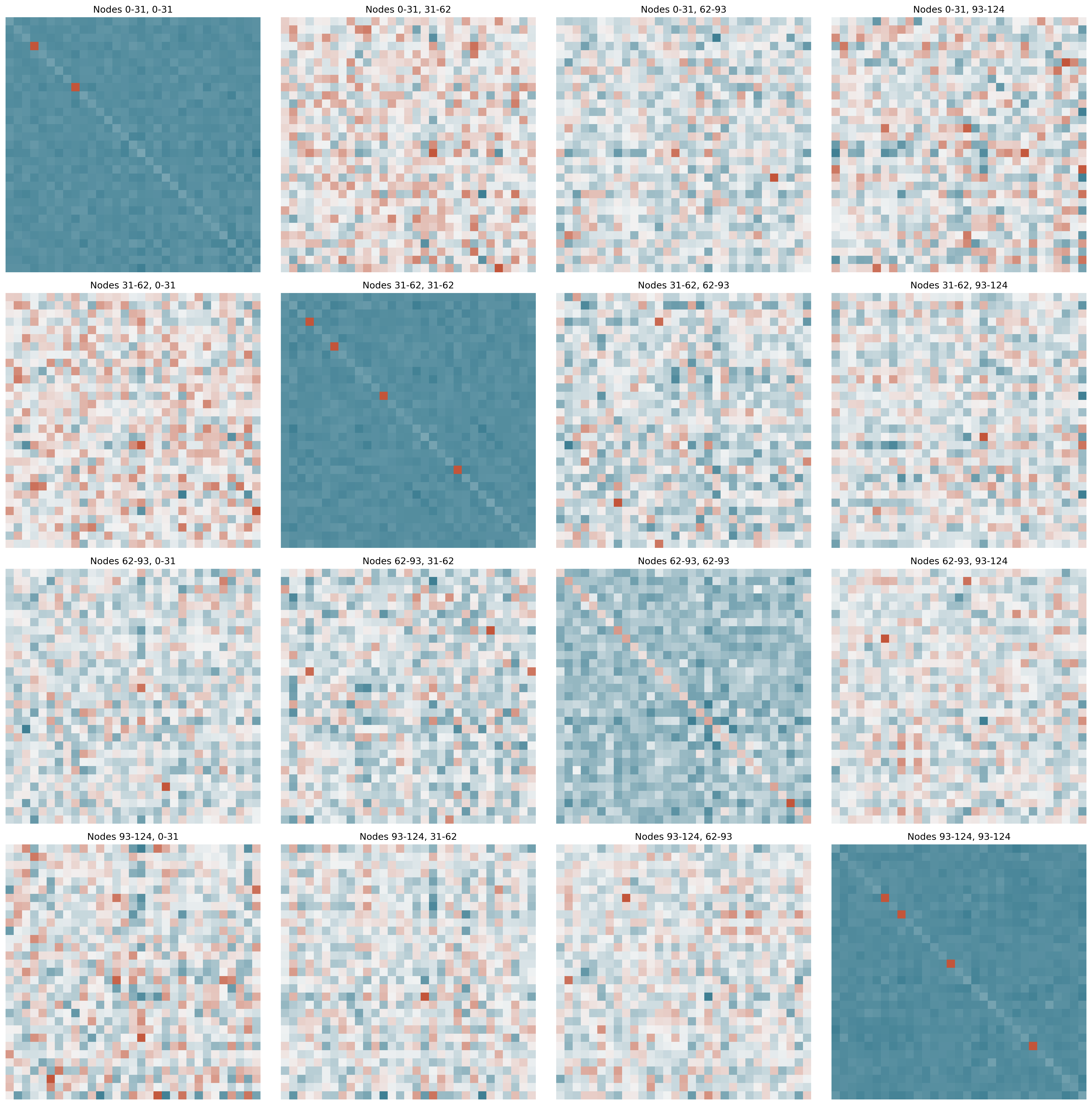} 
        \caption*{Epoch 2}
        \label{fig:Laplacian2}
    \end{minipage}%
    
    \begin{minipage}{0.3\textwidth}
        \centering
        \includegraphics[width=0.95\textwidth]{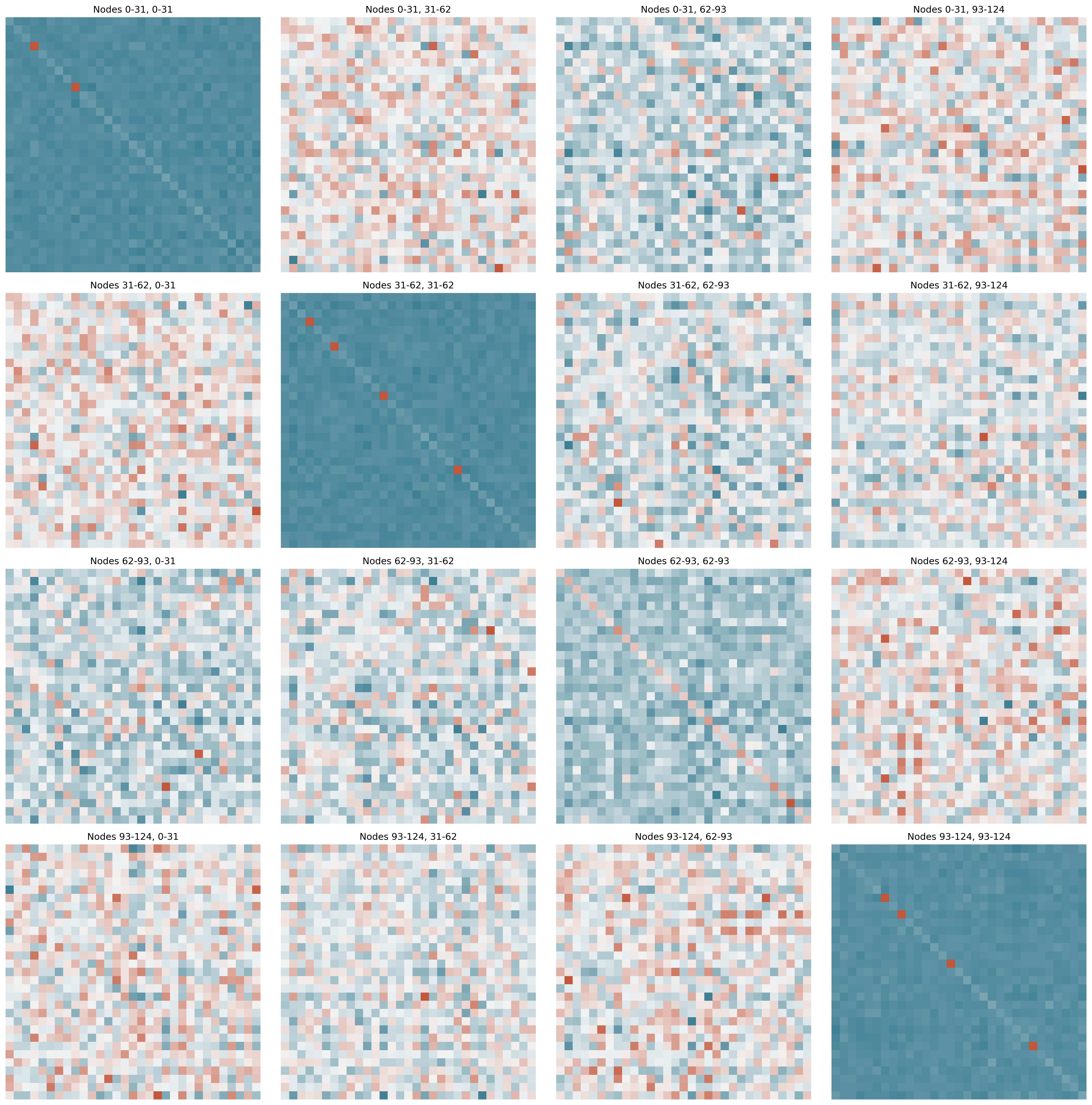} 
        \caption*{Epoch 3}
        \label{fig:Laplacian3}
    \end{minipage}
    
    \vspace{1em} 

    \begin{minipage}{0.3\textwidth}
        \centering
        \includegraphics[width=\textwidth]{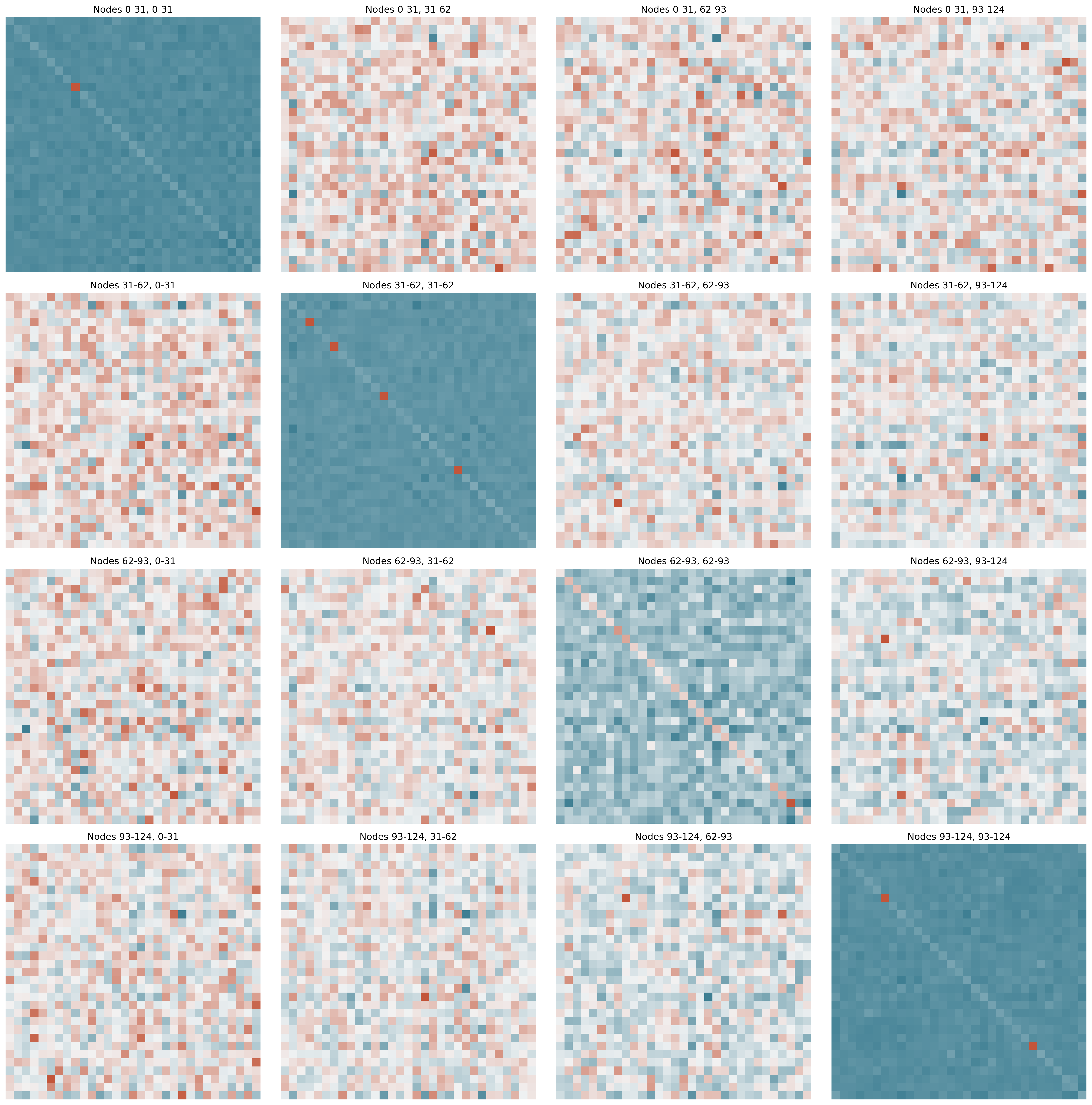} 
        \caption*{Epoch 4}
        \label{fig:Laplacian4}
    \end{minipage}%

    \begin{minipage}{0.3\textwidth}
        \centering
        \includegraphics[width=\textwidth]{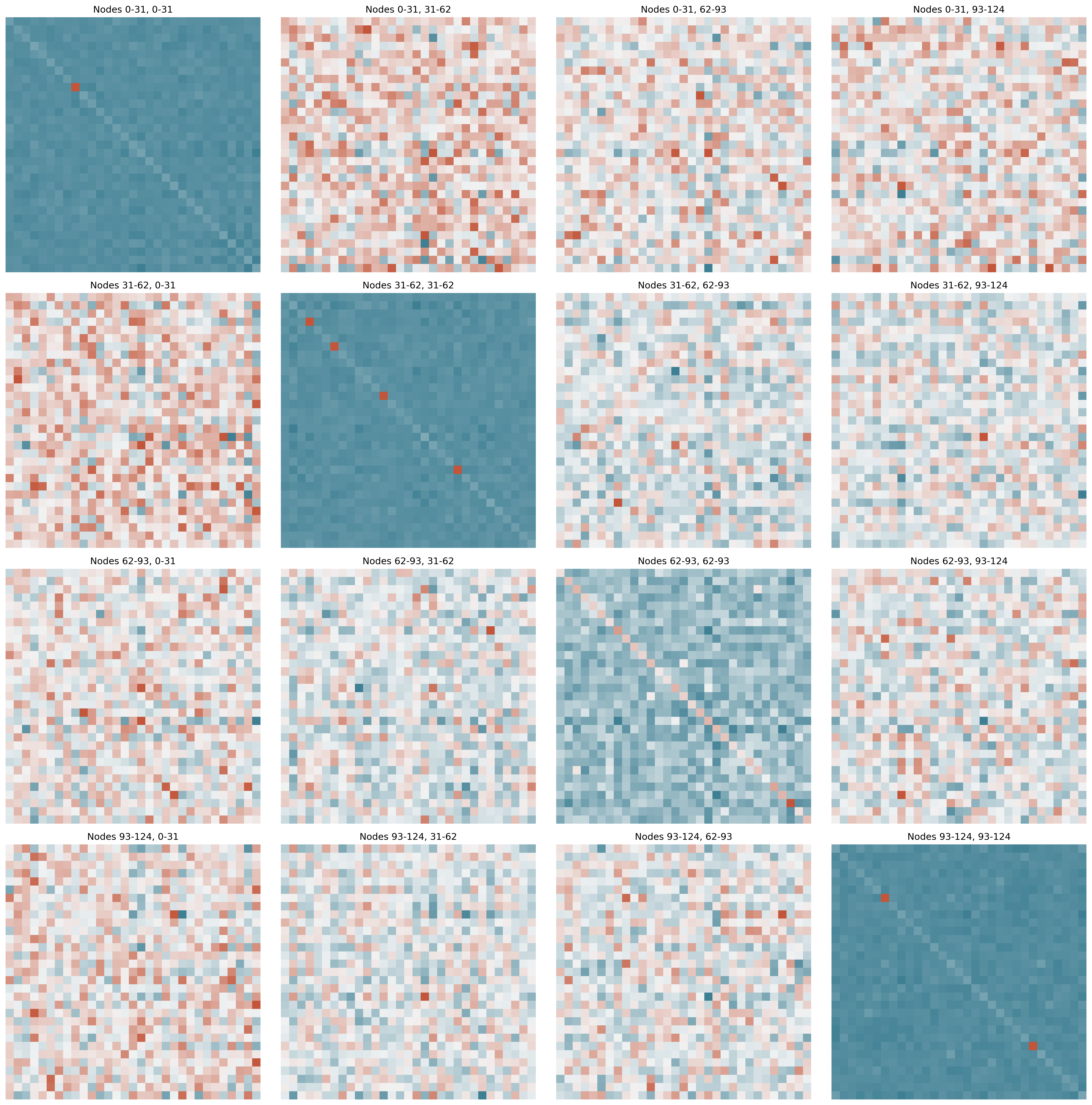} 
        \caption*{Epoch 5}
        \label{fig:Laplacian5}
    \end{minipage}
    } 
    \caption{Feature Correlation Learning Process}
    \caption*{Note: The clusters of red blocks represent feature pairs that are close to positive correlation, the blue blocks represent feature pairs that are close to negative correlation, and the light blocks represent softly correlated feature sets.}
    \label{fig:Epoch_Progression}
\end{figure}

Observing the Figure \ref{fig:Epoch_Progression} , In Epoch 1, the Laplacian matrix presents a relatively random structure, with some weak self-connection signals on the main diagonal. This indicates that the model has just begun to learn the relationship between nodes and has not yet captured obvious structural features. Between Epoch 2-3, it can be observed that the self-connection signal on the main diagonal gradually increases, and local clustering patterns begin to appear in some non-diagonal areas. This shows that the model begins to recognize the association between certain node groups, that is, potential feature clusters. Between Epoch 4-5, the pattern of the non-diagonal area is clearer, showing a block structure, and this structure tends to be stable, indicating that the functional module has successfully learned the dependencies and hierarchical structures between features in the data.

\subsubsection{Spatial-Temporal Convolution Neural Network}
\vspace{1em} 

The Spatial-Temporal Convolution Neural Network (ST-CNN)~\cite{diao2019dynamic} is designed to capture both spatial and temporal dependencies in multivariate time series data. This network combines traditional convolutional neural networks (CNNs) with temporal convolutional networks (TCNs) to effectively model the complex interactions and dynamics present in the data.

\vspace{1em} 

ST-GCN Spatial-Temporal Graph Convolutional Network mainly consists of two parts: GCN and TCN, which are used to capture the temporal dependencies in time series datasets and the dependencies between multiple variables, respectively.

\vspace{1em} 

\textbf{TCN Network Block}:
    The Temporal Convolutional Network (TCN) is primarily used to capture temporal dependencies in time series data. TCN extracts multi-scale temporal features through multi-layer convolution operations and determines the historical length of the time series that each time node can see in the model through the design of the receptive field.

The calculation method of receptive field is:
    \begin{equation}
    \fbox{
        \begin{math}
        \text{receptive\_field} = 
        \begin{cases} 
        1 + (\text{kernel\_size} - 1) \frac{\text{DE}^{\text{layers}-1}}{\text{DE}-1} & \text{if DE} > 1 \\
        \text{layers} \cdot (\text{kernel\_size} - 1) + 1 & \text{if DE} \leq 1
        \end{cases}
        \end{math}
    }
    \end{equation}

    DE means the "dilation\_exponential". The calculation of the receptive field determines the maximum historical length that the model can "see," and its size is determined by the dilation exponential and the number of convolution layers. Specifically:

    \vspace{1em} 
    
    The dilation exponential determines the dilation rate of each convolution layer, which controls the spacing between the elements of the convolution kernel. By adjusting the dilation exponential, the range covered by the convolution kernel can be changed, thereby affecting the size of the receptive field. When the dilation exponential is greater than 1, the spacing of the dilated convolution grows exponentially with the number of layers. This setting allows the receptive field to expand rapidly, enabling the model to see longer historical time series more quickly. When the dilation exponential is less than or equal to 1, the spacing of the dilated convolution remains constant or increases linearly with the layers, causing the receptive field to grow linearly.

    \vspace{1em} 
    
    The number of convolutional layers determines the depth of the network and also affects the size of the receptive field. Increasing the number of convolutional layers can increase the receptive field, allowing the model to capture dependencies over a longer time range. When there are fewer layers, the receptive field is smaller, and the model can only see a shorter historical time series. This is more conducive to capturing short-term dependencies. When the layers are larger, the receptive field also increases, and the model can see a longer historical time series. This allows the model to capture long-term dependencies, but it also increases computational complexity and training difficulty.

\vspace{1em} 

\begin{figure}[h]
\centering
\includegraphics[scale=0.9]{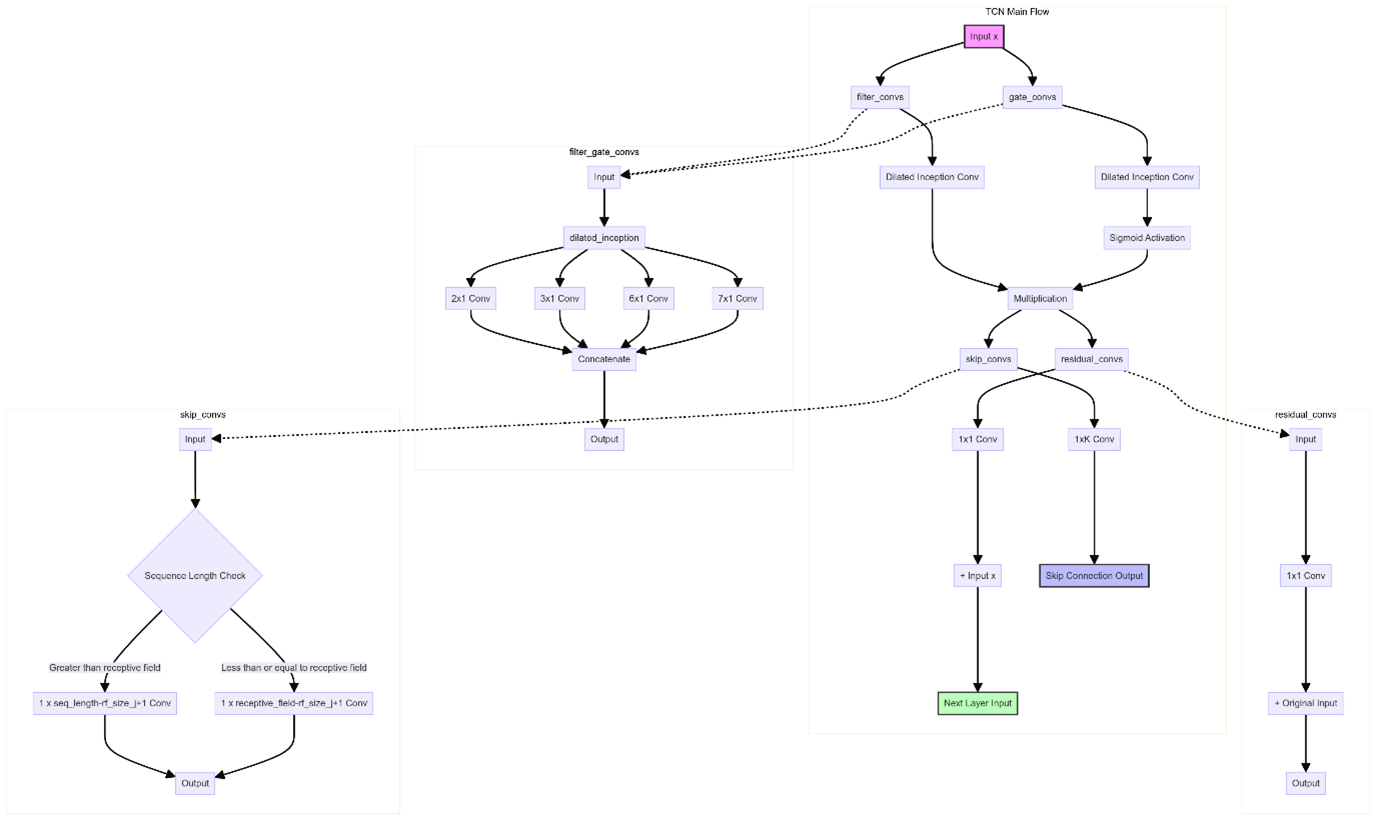} 
\caption{The Workflow for STGCN's TCN Blocks}
\label{fig:TCN_architecture}
\end{figure}

\vspace{1em} 

\textbf{TCN Convolution Block Design}:
The design of the TCN convolution block is crucial for effectively capturing temporal dependencies. The TCN convolution block includes several key components such as filter convolution layers (filter\_convs), gated convolution layers (gate\_convs), residual convolution layers (residual\_convs), and skip convolution layers (skip\_convs), which work together to extract and fuse multi-scale temporal features.

\vspace{1em} 

1. \textbf{Filter Convolutions}: These layers are used to capture multi-scale temporal features, expand the receptive field, and effectively handle long sequence data.

\begin{equation}
\mathbf{F} = \text{dilated\_inception}(X, \text{dilation\_factor} = \text{new\_dilation})
\end{equation}

Filter convolution layers extract multi-scale features from time series by using convolution kernels of different sizes and dilation rates, allowing the model to capture both short-term and long-term dependencies. These convolution layers are constructed based on the dilated\_inception module:

\vspace{1em} 

\begin{itemize}
\item 1.1 \textbf{dilated\_inception} is a key module in TCN used to extract features at different time scales. It achieves this by using convolution kernels of different sizes and dilation rates:

\vspace{1em} 

\item 1.2. \textbf{Multi-scale convolutions}: Using convolution kernels of different sizes (kernel sizes: 2, 3, 6, 7) to extract features at different time scales.

\begin{equation}
X_i = \text{Conv2d}(X, \text{kernel\_size} = k_i, \text{dilation} = \text{dilation\_factor})
\end{equation}

\item 1.3. \textbf{Dilated Convolution}: By introducing intervals in the convolution kernel, each convolution kernel can cover a larger time range, thereby expanding the receptive field. The dilation rate determines the degree of expansion of the convolution kernel.

\begin{equation}
Y_i = \text{Conv2d}(X, \text{kernel\_size} = k_i, \text{dilation} = \text{dilation\_factor})
\end{equation}

\item 1.4. \textbf{Feature stitching}: The output of each convolution kernel is concatenated to form the final output feature.

\begin{equation}
\mathbf{Z} = \text{concat}([\mathbf{Y_1, Y_2, Y_3, Y_4}], \text{dim} = 1)
\end{equation}

\end{itemize}

2. \textbf{Gate Convolutions}:
Gate convolutions are a key component of the Temporal Convolutional Network (TCN), inspired by the gating mechanism~\cite{gu2020improving} in Long Short-Term Memory (LSTM) networks. The primary function of this layer is to dynamically regulate the information flow, thereby enhancing the model's ability to capture complex temporal dependencies in time series data. Gate convolutions work in parallel with the dilated convolution module, both receiving the same input data. 

In a gated convolutional layer, the input is first processed by a 2D convolution operation. This convolution operation usually uses a 1x1 or smaller kernel size, and its purpose is to learn the importance weights of the input features. The output of the convolution operation is then passed through a Sigmoid activation function to compress the value range to between 0 and 1. This step produces a so-called "gate" whose dimensions are consistent with the output of the dilated convolution module (filter). The characteristics of the Sigmoid function make the values close to 1 in the gate indicate that the corresponding features should be retained, while values close to 0 mean that the corresponding features should be suppressed.

The core of the gating mechanism is to perform element-wise multiplication of the gate and the filter. This operation realizes the dynamic selection of features: high values in the gate will retain the corresponding important features in the filter, while low values will suppress less important features. This mechanism gives the network the ability to dynamically select and emphasize important temporal features, and also helps to control the gradient flow and alleviate the gradient vanishing problem in deep networks.

\begin{figure}[H]
    \centering
    \begin{subfigure}[b]{0.3\textwidth}
        \centering
        \includegraphics[width=\textwidth]{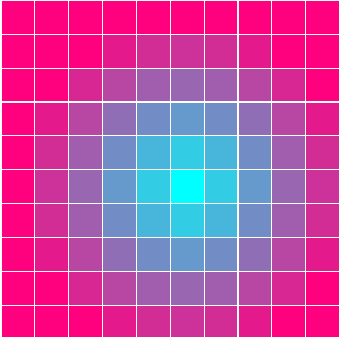} 
        \caption{Gating Signal} 
        \label{fig:subfig1}
    \end{subfigure}
    \hfill
    \begin{subfigure}[b]{0.3\textwidth}
        \centering
        \includegraphics[width=\textwidth]{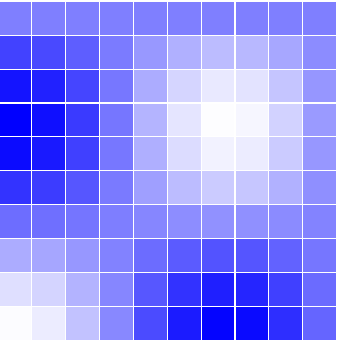} 
        \caption{Input Signal} 
        \label{fig:subfig2}
    \end{subfigure}
    \hfill
    \begin{subfigure}[b]{0.3\textwidth}
        \centering
        \includegraphics[width=\textwidth]{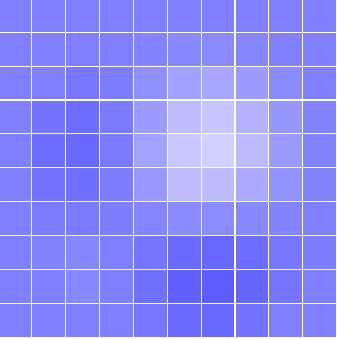} 
        \caption{Output Signal} 
        \label{fig:subfig3}
    \end{subfigure}
    
    \caption{Visualization of gating mechanisms} 
    \caption*{Note: The square matrix in the figure shows the gating form at a certain moment, where the color blocks in the Gating signal are closer to blue, which represents the signal flow that is amplified, while red represents the signal flow that is inhibited. Obviously, the gating at the moment in the figure amplifies the central signal flow.}
    \label{fig:gating}
\end{figure}

Observing Figure \ref{fig:gating}, we can intuitively see how the gating mechanism selectively emphasizes or suppresses information in certain regions. The Gated matrix shows higher gating values at the center (blue) and lower values around the edges (red), exhibiting a pattern that decreases from the center outward (similar to an attention mechanism). The second sub-figure represents the input signal. In the output features after applying the gating mechanism, we can clearly observe that the features in the central region are retained and enhanced, while those in the peripheral regions are significantly suppressed. Through this approach, the gated convolution enhances the model's ability to capture and utilize important patterns in time series data while ignoring irrelevant information, thereby significantly improving the overall modeling performance.

\begin{figure}[H]
\centering
\includegraphics[scale=0.8]{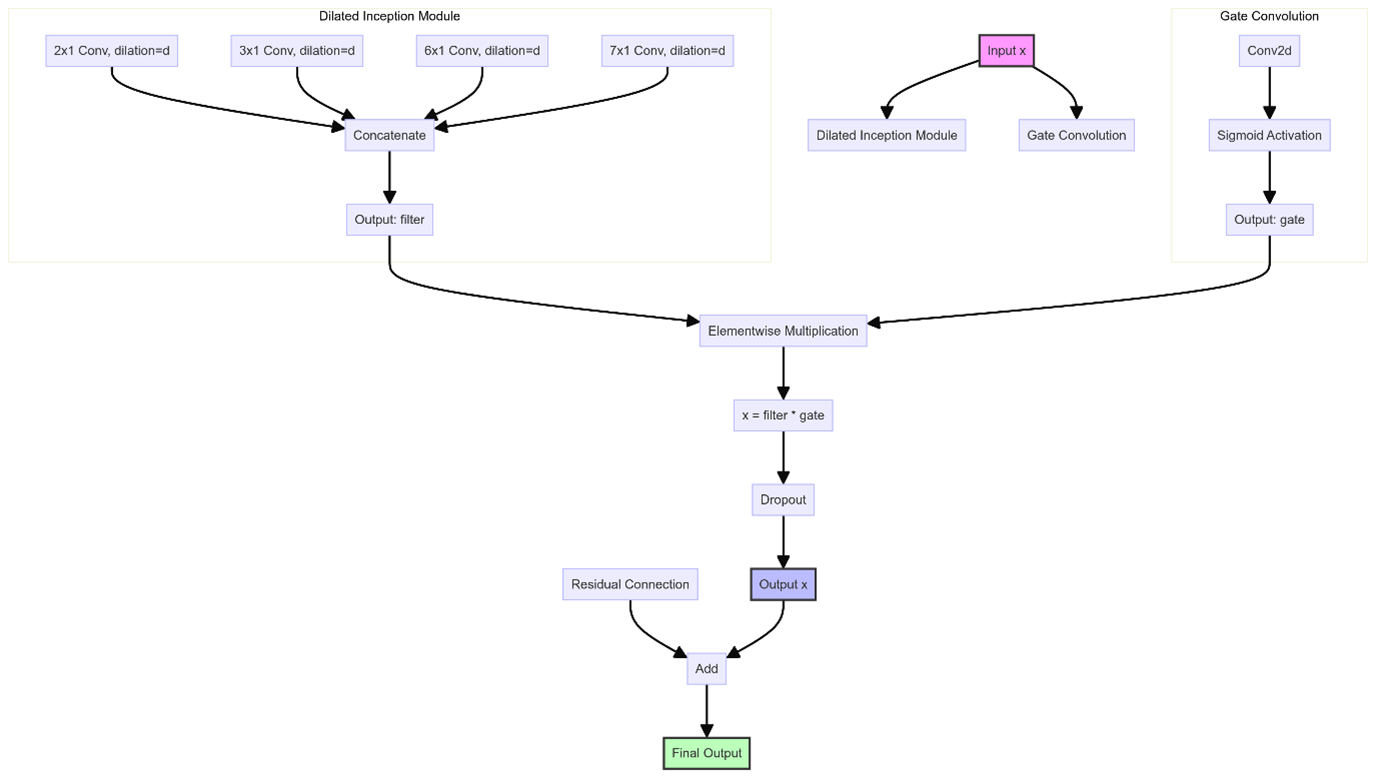} 
\caption{The workflow for gated convolution}
\label{fig:Gated_conv}
\end{figure}

3. \textbf{Residual and Skip Layer}:
The residual convolution layer~\cite{he2016deep} adds a residual connection after the convolution operation, making it easier for the model to learn the identity mapping between input and output, thereby improving the training effect and convergence speed. The skip convolution layer directly transfers the shallow temporal features to the deep layer through the skip connection mechanism~\cite{drozdzal2016importance}, and fuses temporal features of different scales through convolution operations, thereby improving the expressiveness of the model.

\vspace{1em} 

\textbf{GCN Convolution Block Design}:
The Hypergraph Convolutional Block is the core component of the STGCN-Hyper model, designed to capture complex multivariate relationships and higher-order interactions. In my model, the Hypergraph-Conv layer implements message passing and feature aggregation based on the hypergraph structure, effectively handling non-pairwise or multi-hop relationships in spatiotemporal data. The core computation of hypergraph convolution can be expressed as follows:

\begin{equation}
\mathbf{X'} = \sigma (\mathbf{D}_v^{-1/2} \mathbf{H} \mathbf{D}_e^{-1} \mathbf{H}^T \mathbf{D}_v^{-1/2} \mathbf{X} \mathbf{W})
\end{equation}

where $\mathbf{X}$ is the input feature matrix with $N$ nodes and $F$ feature dimensions. $\mathbf{H}$ is the incidence matrix of the hypergraph with $M$ hyperedges. $\mathbf{D}_v = \text{diag}(\mathbf{H1})$ and $\mathbf{D}_e = \text{diag}(\mathbf{1}^T\mathbf{H})$ are the node degree matrix and hyperedge degree matrix, respectively. $\mathbf{W}$ is the learnable weight matrix, and the output feature dimension is $F'$. $\sigma(\cdot)$ is a non-linear activation function.

\subsubsection{Multi-Stage Masking Mechanism}
Based on the STGNN-Hyper model, we introduced an innovative Multi-Stage Masking Mechanism (MSMM) to further enhance the performance of the model in multivariate time series anomaly detection tasks. The core idea of MSMM is to guide the model to learn more robust and general spatiotemporal feature representations by dynamically adjusting the masking strategy of the input data, thereby creating better Node Embeddings for graph nodes. Experiments have shown that this mechanism is suitable for processing complex multivariate time series data, where the data often has high nonlinearity and abnormal fluctuations. The mechanism includes random masking and Laplace-based masking. By dynamically adjusting the mask ratio and the type of mask used, the mechanism encourages the model to learn robust features that can generalize well.

\vspace{1em} 

The main components of the mechanism are as follows:
\begin{itemize}
    \item \textbf{Overall:}
        Let $\mathbf{X} \in \mathbb{R}^{B \times F \times N \times T}$ be the input feature matrix, where $B$ is the batch size, $F$ is the number of features, $N$ is the number of nodes, and $T$ is the sequence length. The `MultiStageMaskingManager` applies a mask $\mathbf{M}$ to the input data, where $\mathbf{M} \in \{0,1\}^{B \times F \times N \times T}$.
    
        The masked input $\mathbf{X'}$ is obtained by the following formula:
        \[
        \mathbf{X'} = \mathbf{X} \circ \mathbf{M}
        \]
    
        where $\circ$ denotes element-wise multiplication.

    \item \textbf{Multi-stage masking strategy:}
        In the Laplacian-based masking stage, the mask is generated based on the importance scores derived from the Laplacian matrix $\mathbf{L}$. The importance score for each node $i$ is given by the diagonal element of $\mathbf{L}$:
        \[
        \text{importance\_score}_i = \mathbf{L}_{ii}
        \]
        
        The masking probability is calculated using the softmax function with temperature parameter $\tau$:
        \[
        p_i = \frac{\exp(\text{importance\_score}_i / \tau)}{\sum_j \exp(\text{importance\_score}_j / \tau)}
        \]
        
        The mask for each node is sampled from the calculated probabilities:
        \[
        \mathbf{M}_i = 
        \begin{cases} 
        1 & \text{if node } i \text{ is selected based on } p_i \\
        0 & \text{otherwise}
        \end{cases}
        \]

    \item \textbf{Time decay factor:}
        Additionally, a temporal decay factor $\alpha$ is introduced to adjust the masking strength at different time steps:
        \[
        w(t) = \frac{\alpha^t}{\sum_{k=0}^{\text{seq\_length}-1} \alpha^k}
        \]
        
        This ensures that the most recent time steps receive less masking, while more distant time steps receive more masking, encouraging the model to focus more on recent observations. This is consistent with practical scenarios where it is more likely to lose feature data from distant time steps.
\end{itemize}

\subsubsection{Graph Structure Learning}
Based on the STGNN-Hyper model, we introduce an innovative Graph Structure Learning (GSL) module~\cite{jin2020graph} to dynamically learn and optimize the relationships between feature nodes in multivariate time series data. 
The core idea of the GSL module is to adaptively construct the graph structure by learning node embeddings, rather than relying on predefined static graphs. The working principle is as follows:

\begin{itemize}
    \item \textbf{Node Embedding:} Each node is represented by an embedding vector, which is initialized randomly during the training process. This is implemented using \texttt{nn.Embedding}.

    \item \textbf{Residual Correlation Calculation:} A residual correlation calculation is used to measure the correlation between node embeddings. Given the embedding matrix $\mathbf{E} \in \mathbb{R}^{N \times d}$, where $N$ is the number of nodes and $d$ is the embedding dimension, the residual correlation matrix $\mathbf{C} \in \mathbb{R}^{N \times N}$ is calculated as:
    \[
    C_{ij} = \frac{\mathbf{E}_i \cdot \mathbf{E}_j}{\|\mathbf{E}_i\| \|\mathbf{E}_j\|}
    \]
    Here, $\mathbf{E}_i$ and $\mathbf{E}_j$ represent the embeddings of nodes $i$ and $j$.

    \item \textbf{Top-K Selection:} For each node, the $K$ most relevant nodes are selected based on the residual correlation. This forms a neighborhood graph used for the subsequent GCN operations. The top-$K$ indices for node $i$ are represented as:
    \[
    \text{topk\_indices}_{ij} = \text{argsort}(C_{ij})[:K]
    \]

    \item \textbf{Graph Construction:} The learned graph structure is represented as an adjacency matrix $\mathbf{R} \in \mathbb{R}^{2 \times (N \times K)}$, where each row represents the indices of connected nodes. The adjacency matrix is then used to guide the message-passing process~\cite{liu2021interest} in the GCN, facilitating effective feature learning.
\end{itemize}

In the STGNN-Hyper model, the GSL module works in conjunction with the Temporal Convolutional Network (TCN) and Graph Convolutional Network (GCN):
\begin{itemize}
    \item \textbf{TCN captures temporal dynamics for each node:} 
    \[
    \mathbf{X}_t = f_{\text{TCN}}(\mathbf{X}_{t-1}, \ldots, \mathbf{X}_{t-r})
    \]
    where $r$ is the receptive field size.

    \item \textbf{GSL learns and updates the graph structure:} 
    \[
    \mathbf{A}_t = \text{GSL}(\mathbf{X}_t)
    \]
    $\mathbf{A}_t$ is the adjacency matrix at time step $t$.

    \item \textbf{GCN propagates spatial information on the learned graph:} 
    \[
    \mathbf{H}_t = \sigma(\mathbf{D}^{-1/2}\mathbf{A}_t\mathbf{D}^{-1/2}\mathbf{X}_t\mathbf{W})
    \]
    where $\mathbf{D}$ is the degree matrix, $\mathbf{W}$ is the learnable weight matrix, and $\sigma$ is the non-linear activation function.
\end{itemize}

\subsubsection{MLP Predictor}
\vspace{1em} 

Finally, the model obtains the final prediction result by fusing the output of GCN into a multi-layer perceptron MLP~\cite{taud2018multilayer}. This fusion allows the model to fully utilize the complex spatial relationships captured by the hypergraph structure while performing nonlinear feature transformation and prediction through MLP.

\begin{equation}
\hat{\mathbf{Y}} = f_{\text{MLP}}(f_{\text{GCN}}(\mathbf{X}, \mathbf{L}))
\end{equation}

Here, $\mathbf{H}_0$ denotes the input features after hypergraph convolution, $\mathbf{W}_i$ and $\mathbf{b}_i$ represent the weight matrix and bias vector of the $i$-th layer, respectively, $\sigma$ represents the activation function (e.g., ReLU), and $k$ represents the number of layers in the MLP.

\begin{equation}
h_0 = Z
\end{equation}

\begin{equation}
h_i = \text{ReLU}(\text{BN}(W_i h_{i-1} + b_i)), \quad i = 1, \ldots, L-1
\end{equation}

\begin{equation}
\hat{\mathbf{Y}} = W_L h_{L-1} + b_L
\end{equation}

Here, $L$ is the number of layers in the MLP, $W_i$ and $b_i$ are the weight matrix and bias vector of the $i$-th layer, respectively, BN represents the batch normalization operation~\cite{santurkar2018does}, and ReLU is the rectified linear unit activation function. The final layer $W_L$ and $b_L$ are used to generate the final prediction output.

\subsection{Anomaly Detector}
\begin{figure}[H]
\centering
\includegraphics[width=\textwidth]{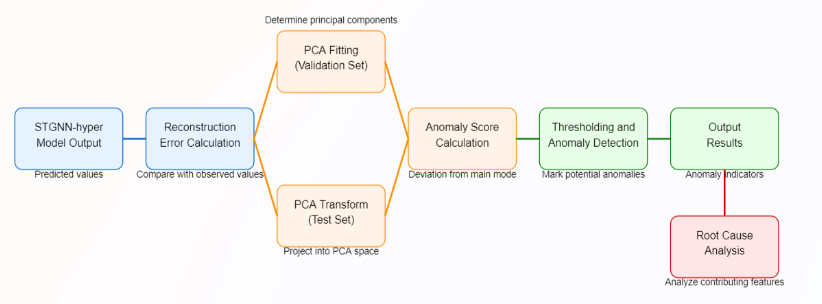} 
\caption{PCA Detector Workflow}
\label{fig:PCA_workflow}
\end{figure}

\hspace{2em}\textbf{Method 1 (PCA)} - Observing Figure \ref{fig:PCA_workflow}, The output of the STGNN-hyper model is the predicted value for each time step, which we compare with the actual observed value and calculate the reconstruction error. The reconstruction error is input into the PCA module for anomaly detection. PCA~\cite{abdi2010principal} performs a PCA transformation on the reconstruction error of the validation set to determine the number of principal components. We then project the reconstruction error of the test set into the same PCA space. By calculating the difference between the original reconstruction error and its PCA reconstruction, we can quantify the degree to which each error value deviates from the main mode. This deviation is used as an anomaly score, and data points above a certain threshold are marked as potential anomalies. In addition, by analyzing the principal components that contribute most to the anomaly, we can trace back to the original feature space to find the root features that cause the anomaly. A better diagnostic method will be proposed later in this paper.

The algorithm includes the following main steps:
\begin{itemize}
    \item Using the STGNN-Hyper model to make predictions on the validation and test sets.
    \item Computing prediction errors and normalizing them.
    \item Fitting a PCA model using the validation set data.
    \item Transforming the target data of the test set into the PCA space and reconstructing it.
    \item Computing the reconstruction errors and using them as anomaly scores.
    \item Setting a threshold and generating anomaly predictions (lines 26-29).
\end{itemize}

Additionally, there are two auxiliary functions: \textit{ErrorNormalizer} and \textit{SlidingWindowNormalizer}, which are used for normalizing the sliding window data.

\vspace{1em} 

\hspace{1em}\textbf{Method 2 (GMM)} - Similarly, the output of the previous model is the predicted value for each time step, which we compare with the actual observed value to calculate the reconstruction error. The reconstruction error is then fed into the GMM module for anomaly detection. GMM~\cite{reynolds2009gaussian} assumes that the data consists of multiple Gaussian distributions, each representing a sub-mode of the data. The reconstruction error is first normalized to eliminate the scale differences between different features. Then, the GMM model is fitted using the validation set data. The number of components of the model is a key parameter, which is determined by optimizing the F1 score on the validation set. The fitted GMM model is used to calculate the log-likelihood of the validation and test set data. Normal data points have high log-likelihood values, while anomalies have low log-likelihood values. Therefore, we use the negative log-likelihood as the anomaly score. By setting a threshold on the validation set (usually the maximum anomaly score is chosen), we can mark the data points in the test set whose anomaly score exceeds this threshold as anomalies. GMM is able to capture the complexity of the data distribution and is not limited to a single Gaussian assumption. It can accommodate multiple normal modes.

\section{Experiments Study}
In this section, we first introduce the datasets used in the experiments and explain the performance indicators used. First, in our baseline comparison experiments, we compare the model STGCN-Hyper with the most advanced deep learning models on multiple datasets. Then, the necessity of each component of our model is verified through ablation experiments. Similarly, we compare the detection effects of anomaly detectors based on different principles and verify the rationality of the assumptions they are based on. Finally, we experimentally verify the interpretability of anomalies provided by our model and its role in helping anomaly diagnosis tasks.

\subsection{Datasets Descriptions}

\begin{table}[h]
\centering
\caption{Dataset Information}
\vspace{0.5cm} 
\begin{tabular}{lcccc}
\toprule
\textbf{Dataset} & \textbf{Dimensions} & \textbf{Train} & \textbf{Test} & \textbf{Anomalies (\%)} \\
\midrule
SMD & 38 & 708420 & 708420 & 4.16 \\
WADI & 127 & 118795 & 17275 & 5.99 \\
SWaT & 51 & 47515 & 44986 & 11.97 \\
SMAP & 25 & 135183 & 427617 & 13.13 \\
MBA & 2 & 100000 & 100000 & 0.14 \\
\bottomrule
\end{tabular}
\end{table}

\begin{itemize}
    \item \textbf{SMD (Server Machine Dataset)} SMD (Server Machine Dataset) is a new 5-week dataset. The dataset contains 3 groups of entities. Each group is named machine-<group-index>-<index>. SMD consists of data from 28 different machines, and 28 subsets should be trained and tested separately. For each subset, split it into two parts with equal lengths for training and testing separately. The dataset provides a label for whether a point is an anomaly or not, and the contribution to each dimension of the anomaly.

    \item \textbf{WADI Dataset} This dataset collects all the data generated by 127 devices running continuously for 16 days (2 days in 15 attack scenarios and the rest of the days in normal operation). In the training data, WADI only contains normal data. WADI is 127-dimensional high-dimensional data and the data is unbalanced.

    \item \textbf{SWaT (Secure Water Treatment Dataset)} The dataset contains data from 11 consecutive days of operation: 7 days of normal operation and 4 days of attack scenarios. Network traffic and all values obtained from all 51 sensors and actuators are collected and labeled according to normal and abnormal behavior. The dataset records a series of malware infection attacks against SWaT engineering workstations. Malware attacks include historical data exfiltration attacks and process disruption attacks.

    \item \textbf{SMAP (Soil Moisture Active Passive Dataset)} This dataset is generated by integrating satellite-derived Soil Moisture Active Passive (SMAP) Level 2 soil moisture observations into a modified two-layer Palmer model using a one-dimensional ensemble Kalman filter (EnKF)~\cite{simon2001kalman} data assimilation method. Soil moisture anomalies are calculated based on the climatological conditions on the target day. The climatological conditions are estimated based on the full data record of SMAP satellite observations and a moving window method centered around 31 days. The assimilation of SMAP soil moisture observations helps improve model-based soil moisture predictions, especially in areas with poor instrumentation where good quality precipitation data is lacking.

    \item \textbf{MBA (MIT-BIH Supraventricular Arrhythmia Datasat)} This database includes 78 half-hour ECG records and is used to supplement the sample of supraventricular arrhythmias in the MIT-BIH Arrhythmia Database.
\end{itemize}

\subsection{Anomaly Detection Performance}
In this section, we will use the two indicators of Precision and Recall to test the performance of the STGCN-based model on high-dimensional time series datasets.

\begin{table}[H]
\centering
\scriptsize
\begin{tabular}{|c|c|c|c|c|c|c|c|c|}
\hline
\multirow{2}{*}{} & \multicolumn{2}{c|}{SWaT} & \multicolumn{2}{c|}{WADI} & \multicolumn{2}{c|}{SMD} \\
\cline{2-7}
 & Pre & Rec & Pre & Rec & Pre & Rec \\
\hline
LSTM-NDT & 0.7600 & 0.8950 & 0.1117 & 0.6544 & 0.7855 & 0.8528 \\

DCdetector & 0.8257 & 0.7298 & 0.5597 & 0.2731 & 0.6742 & 0.2189 \\

OmniAnomaly & 0.8142 & 0.7224 & 0.3160 & 0.6540 & 0.8880 & 0.9960 \\

LOF & 0.7215 & 0.6543 & 0.1158 & 0.0900 & 0.5634 & 0.3986 \\

OCSVM & 0.4539 & 0.4922 & 0.2435 & 0.2401 & 0.4434 & 0.7672 \\

MAD-GAN & 0.6401 & 0.8746 & 0.2230 & 0.7310 & 0.7290 & 0.8409 \\

DAGMM & 0.8992 & 0.5784 & 0.2510 & 0.8290 & 0.5951 & 0.8782 \\

GDN & 0.8124 & 0.6812 & 0.2910 & 0.7930 & 0.8443 & 0.4684 \\

\hline
STGCN-Hyper & \textbf{0.8520} & \textbf{0.7628} & \textbf{0.8736} & \textbf{0.5751} & \textbf{0.9420} & \textbf{0.9970} \\

\hline
\end{tabular}

\vspace{10pt}  
\caption{Performance of different models on high-dimensional datasets}
\label{tab:results}
\end{table}

The STGCN-Hyper model demonstrated outstanding performance on both the SWaT and WADI datasets. In the SWaT dataset, the model achieved an accuracy of 0.8520 and a recall of 0.7628, significantly outperforming other baseline models.

Notably, the improvement of the STGCN-Hyper model was most evident on the WADI dataset. It achieved an accuracy as high as 0.8736, far surpassing other models, such as DAGMM (accuracy 0.2510) and GDN (accuracy 0.2910). Although the recall (0.5751) did not improve substantially, it was still superior to that of other models.

Compared to other models, STGCN-Hyper also excelled in the balanced metric F1-score:

\begin{itemize}
\item On the SWaT dataset, STGCN-Hyper's performance comprehensively surpassed traditional methods like LOF~\cite{breunig2000lof} (accuracy 0.7215, recall 0.6543) and OCSVM (accuracy 0.4539, recall 0.4922).

\item Compared to more recent deep learning methods, such as OmniAnomaly (accuracy 0.8142, recall 0.7224 on SWaT) and GDN (accuracy 0.8124, recall 0.6812 on SWaT), STGCN-Hyper showed an improvement in balancing accuracy and recall.

\item The advantages of STGCN-Hyper were even more pronounced on the WADI dataset. Other models, such as LSTM-NDT (accuracy 0.1117, recall 0.6544) and MAD-GAN (accuracy 0.2230, recall 0.7310), showed that while the MAD model had a relatively high recall, its accuracy was low. In contrast, STGCN-Hyper maintained a high recall while also ensuring accuracy, preventing a large number of false positives.
\end{itemize}

\subsection{Visualization of Prediction Performance}
\begin{figure}
\centering
\resizebox{0.85\textwidth}{!}{ 
    \begin{minipage}{\textwidth}
        \centering
        \begin{subfigure}{0.45\textwidth}
            \centering
            \includegraphics[width=\textwidth]{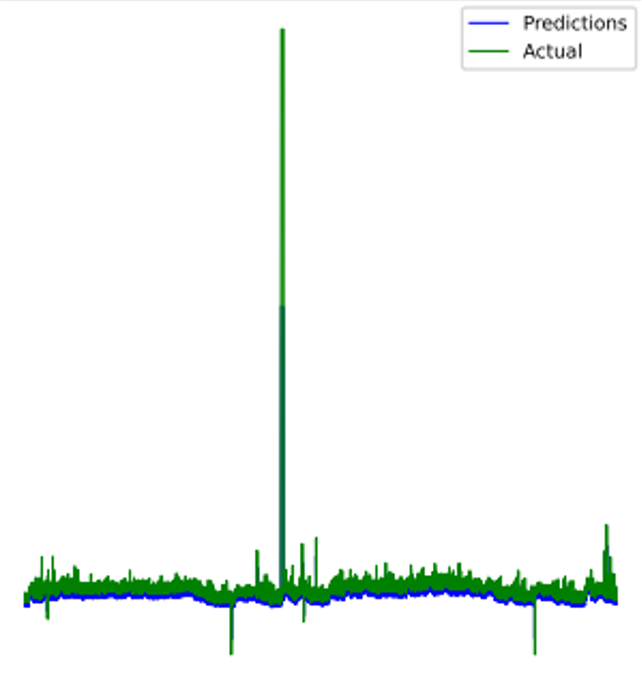} 
            \caption{Stationary Sequence}
            \label{fig:v1}
        \end{subfigure}
        \hfill
        \begin{subfigure}{0.45\textwidth}
            \centering
            \includegraphics[width=\textwidth]{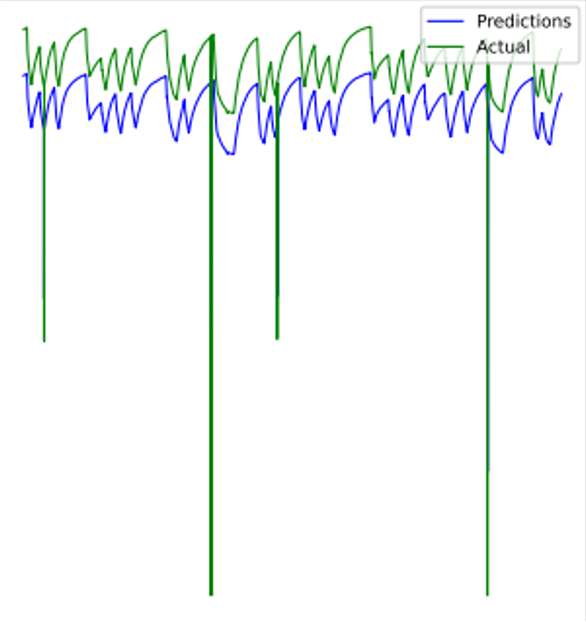} 
            \caption{Periodic Fluctuation Sequence}
            \label{fig:v2}
        \end{subfigure}

        \vspace{0.5cm}

        \begin{subfigure}{0.45\textwidth}
            \centering
            \includegraphics[width=\textwidth]{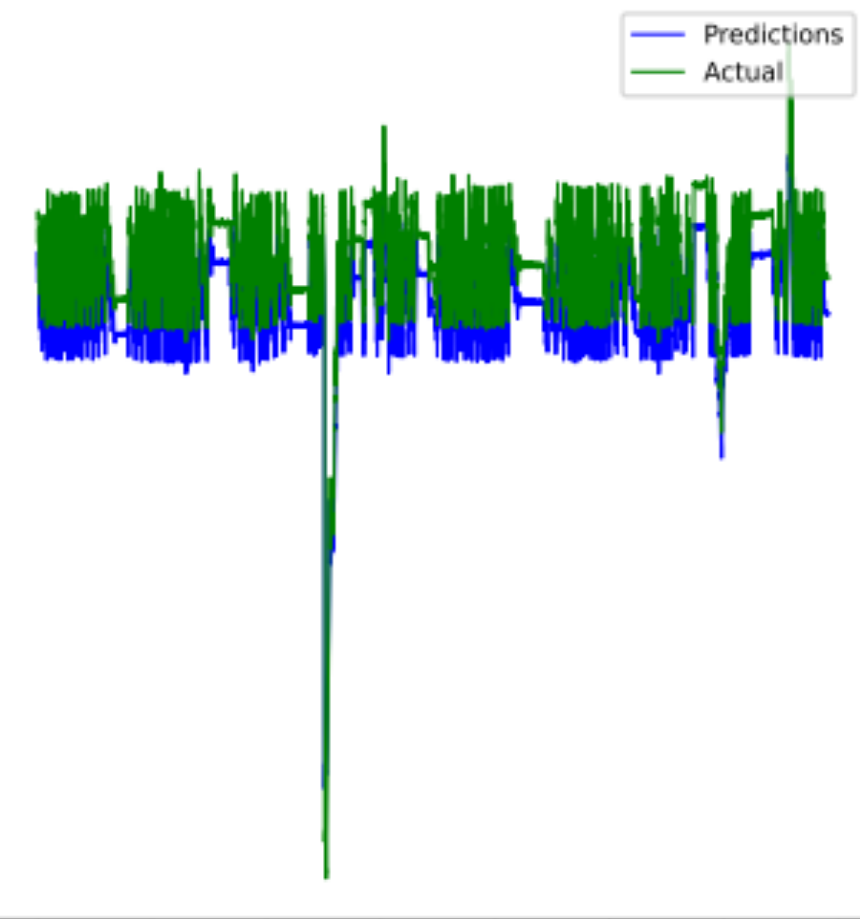} 
            \caption{Fluctuation Sequence}
            \label{fig:v3}
        \end{subfigure}
        \hfill
        \begin{subfigure}{0.45\textwidth}
            \centering
            \includegraphics[width=\textwidth]{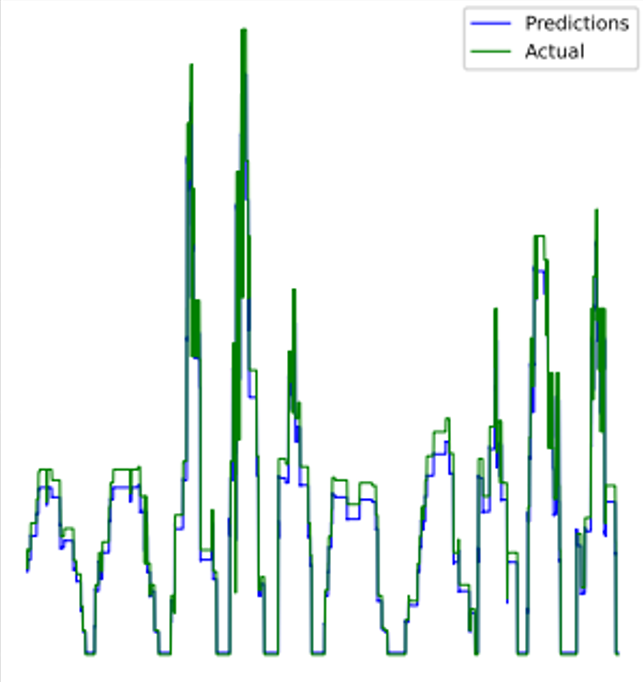} 
            \caption{Seasonal Spike Sequence}
            \label{fig:v4}
        \end{subfigure}

        \vspace{0.5cm}

        \begin{subfigure}{0.45\textwidth}
            \centering
            \includegraphics[width=\textwidth]{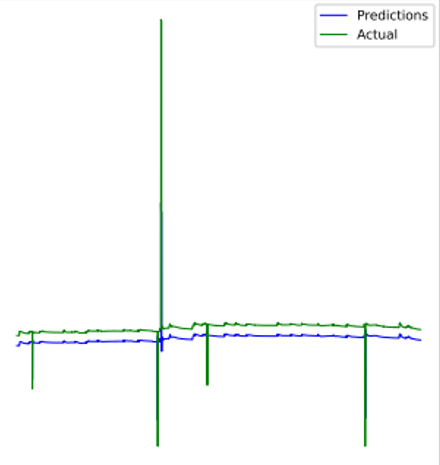} 
            \caption{Step Change Sequence}
            \label{fig:v5}
        \end{subfigure}
        \hfill
        \begin{subfigure}{0.45\textwidth}
            \centering
            \includegraphics[width=\textwidth]{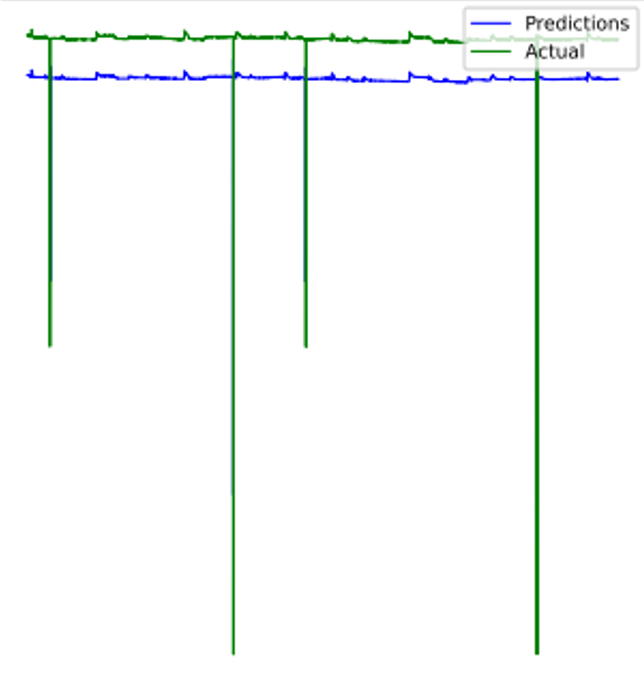} 
            \caption{Step Change Sequence2}
            \label{fig:v6}
        \end{subfigure}
    \end{minipage}
}
\vspace{10pt}
\caption{Various features actual and predicted curves}
\label{fig:2x3grid}
\end{figure}

The stgcn-hyper model can perform time series forecasting and anomaly detection in multivariate time series datasets in many different fields. In this paper, we take the WADI dataset as an example, and randomly select six representative variables from it, use their actual value curves and predicted curves for visualization analysis, and observe the performance characteristics of the model and its ability in anomaly detection tasks:

\begin{enumerate}
    \item \textbf{Stable Stationary Sequence (Figure \ref{fig:v1})}
    The sequence remains at a stable level between 0.2 and 0.4 for most of the time. The model (blue line) accurately captures this stationary characteristic, with predicted values closely following the actual values (green line). Notably, the sequence contains several significant spike anomalies, with the highest reaching around 3.0. The model's predictions are able to detect these anomalous changes, maintaining the original trend of data changes, thus providing the capability to detect these anomalous data points.
    
    \item \textbf{Periodic Fluctuation Sequence (Figure \ref{fig:v2})} 
    This sequence exhibits a clear periodicity, with fluctuations ranging approximately between 0.6 and 1.0. The model accurately learns and predicts this periodic pattern, with the predicted curve nearly matching the amplitude of the actual curve. Additionally, the sequence includes four sharp drops (down to near 0 or 0.4). The model successfully captures these anomalies, indicating that the model can detect not only high-value anomalies beyond the normal range but also low-value anomalies that deviate from the periodic pattern, demonstrating its anomaly detection capability in periodic data.
    
    \item \textbf{High-Frequency Fluctuation Sequence (Figure \ref{fig:v3})} 
    This sequence shows high-frequency, intense fluctuations, including multiple extreme anomalies. While the model's predictions (blue line) do not perfectly match every high-frequency fluctuation, they form a trend that closely follows the fluctuations, with predicted values slightly lower than the actual values. This reflects the model's ability to fit high-frequency fluctuation trends. The model successfully identifies extreme anomalies within the sequence, particularly the large negative anomaly. This performance suggests that the model can maintain an overall trend while not overlooking significant anomalous signals.
    
    \item \textbf{Irregular Seasonal Spike Sequence (Figure \ref{fig:v4})} 
    This sequence presents an irregular spike pattern with a constantly changing baseline level. The model quickly adjusts its predictions to match the changing baseline. For most seasonal spikes, the model makes timely predictions, although in some cases the predicted magnitude may be smaller than the actual value. This performance reflects the model's ability to handle non-stationary time series, remaining sensitive to sudden events without overreacting to every seasonal fluctuation.
    
    \item \textbf{Step Change Sequence (Figure \ref{fig:v5})}
    This sequence displays significant step changes in level, abruptly rising from about 0.7 to nearly 1.0. The model performs excellently when faced with such sharp structural changes. It recognizes the anomalous step-like data changes, with little influence from outliers, and the prediction curve remains within the range of normal data variations. At the same time, the model maintains a good fit for small fluctuations within the stationary parts of the sequence.
    
\end{enumerate}

\subsection{Ablation study for the STGCN-Hyper model}
To comprehensively evaluate the contributions of each component in the STGNN-hyper model for anomaly detection tasks, we conducted in-depth ablation experiments on an industrial control system multivariate time series dataset: WADI. We implemented several variants of the STGNN-hyper model, systematically modifying different modules to quantify the impact of each component:

\begin{itemize}
    \item \textbf{Without Hypergraph (w/o Hypergraph):} In this variant, we replaced the hypergraph structure with a traditional graph structure. Specifically, we replaced the hypergraph adjacency matrix \( \mathbf{H} \) with a standard binary graph adjacency matrix \( \mathbf{A} \) and substituted the hypergraph convolution operation with a conventional 2D graph convolution. This modification aims to assess the advantage of the hypergraph structure in capturing high-order variable relationships.
    \item \textbf{Without TCN (w/o TCN):} We removed the Temporal Convolutional Network (TCN) module and instead used a simple Multi-Layer Perceptron (MLP) to handle temporal information. This change allows us to quantify and evaluate the effect of the TCN network in capturing multi-scale temporal patterns, i.e., features in both short-term and long-term temporal variations.
    \item \textbf{Without GCN (w/o GCN):} In this setup, we removed the Graph Convolutional Network (GCN) module, including the corresponding hypergraph convolution operations. We replaced the GCN with a fully connected linear layer to process spatial information. This modification helps us understand the importance of GCN in aggregating node features and propagating spatial information, and evaluate the role and accuracy of the hypergraph neural network in capturing dynamic interdependencies between variables.
    \item \textbf{Without MTCL (w/o MTCL):} We removed the Multivariate Time-series Correlation Learning module (hypergraph\_constructor\_mu) and instead used a fixed, predefined hypergraph structure. Specifically, we used a fixed identity matrix to represent static associations between feature nodes. This change aims to assess the role of the adaptive Laplacian matrix learning module in capturing dynamic associations between variables.
    \item \textbf{Full model:} The complete STGNN-hyper model, including all components, serves as the baseline to evaluate other variants.
\end{itemize}

Table 3 presents the detailed ablation experiment results. We use Precision, Recall and F1 Score as the evaluation metrics:

\begin{table}[h]
\centering
\begin{tabular}{|c|c|c|}
\hline
       & Precision   & Recall   \\ \hline
Full model & 0.8520 & 0.7628 \\ \hline
w/o Hyper-Structure & 0.3082 & 0.6065 \\ \hline
w/o TCN  & 0.6307 & 0.4087 \\ \hline
w/o GCN & 0.5313 & 0.5601 \\ \hline
w/o MTCL  & 0.8750 & 0.3976 \\ \hline
\end{tabular}
\vspace{10pt}
\caption{Ablation experiment results on WADI datasets.}
\end{table}

Based on the test results in Table 3, we can make the following analyses:
\begin{enumerate}
    \item \textbf{Superiority of the complete model:} The complete STGNN-hyper model outperforms all ablation variants on all evaluation metrics, achieving the highest precision (0.8520) and recall (0.7628). This confirms the synergy between components, enabling the model to effectively capture complex spatiotemporal dependencies and anomaly patterns.
    \item \textbf{Key role of the hypergraph structure:} Replacing the hypergraph structure with a traditional binary graph structure results in significant performance degradation, with precision dropping from 0.8520 to 0.4155 (-51.2\%) and recall dropping from 0.7628 to 0.3976 (-47.8\%). This significant decrease highlights the advantage of the hypergraph in modeling high-order relationships between variables in complex systems. The hypergraph structure can simultaneously capture complex interactions among multiple variables, which is difficult to achieve in traditional binary graphs. The hypergraph structure shows a significant advantage in improving detection precision, likely due to its ability to more accurately describe complex dependencies in the system, thereby reducing false positives. It also shows a significant advantage in improving detection recall, as hypergraph convolution can capture associations between multi-hop neighboring nodes, thus more comprehensively detecting system anomalies.
    \item \textbf{Importance of temporal modeling:} Replacing the TCN module with an MLP results in a performance decline, especially in precision (from 0.8520 to 0.6307, -26.0\%). The decrease in recall is relatively smaller (from 0.7628 to 0.6756, -11.4\%). This indicates that TCN has significant advantages in capturing multi-scale temporal patterns. TCN can handle both short-term fluctuations and long-term trends, which is crucial for accurately identifying anomaly patterns because each node in the node embedding processed by TCN can ensure its receptive field covers the entire detection period. While MLP can also handle temporal information, it is not as flexible as TCN in capturing patterns at different time scales, explaining why its impact on precision is greater than on recall.
    \item \textbf{Value of spatial information propagation:} Replacing the GCN module with a fully connected linear layer results in significant performance degradation, with precision decreasing by 37.6\% (from 0.8520 to 0.5313) and recall decreasing by 26.6\% (from 0.7628 to 0.5601). This emphasizes the critical role of GCN in aggregating node features and propagating spatial information. GCN can effectively capture dynamic interdependencies between variables, which is crucial for understanding the overall state of the system and detecting anomalies. Since the information propagation of GCN is based on the Laplacian matrix learned by the preceding model, it allows each node to more effectively aggregate features of related nodes, reducing the aggregation of irrelevant node information and enhancing the purity of the node embedding. Although a fully connected layer can also process information from all nodes, it lacks the structured information propagation capability of GCN, leading to significant performance decline.
    \item \textbf{Core position of adaptive correlation learning:} Replacing the MTCL module with a fixed identity matrix has the most significant negative impact on recall, dropping from 0.7628 to 0.3976 (-47.9\%). However, precision slightly increases (from 0.8520 to 0.8750, +2.7\%). This interesting phenomenon suggests that the MTCL module plays a crucial role in capturing various anomaly patterns, especially in identifying complex or subtle anomalies. A fixed correlation structure may miss dynamic changes in the system, leading to a large number of anomalies being undetected (false negatives), explaining the significant drop in recall. The slight increase in precision may be because the fixed structure reduces some noise effects, but this "stability" comes at the cost of the model's sensitivity to anomalies.
\end{enumerate}

\subsection{Anomaly Detectors Comparisons}
\subsubsection{PCA}
\begin{figure}[H]
\centering
\raisebox{-0.5\height}{\begin{minipage}{0.32\textwidth}
  \centering
  \includegraphics[width=\textwidth]{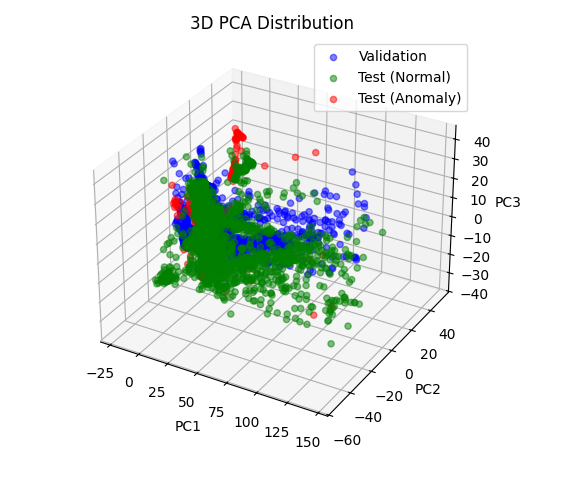} 
  \caption{Original Model}
  \label{fig:pca_re1}
\end{minipage}}
\hfill
\raisebox{-0.5\height}{\begin{minipage}{0.32\textwidth}
  \centering
  \includegraphics[width=\textwidth]{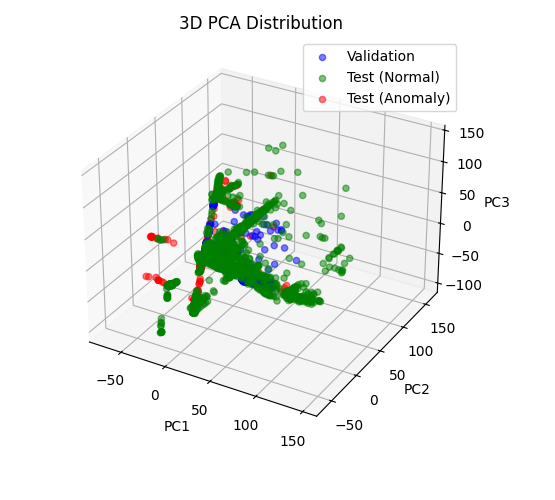} 
  \caption{Training With Masking}
  \label{fig:pca_re2}
\end{minipage}}
\hfill
\raisebox{-0.5\height}{\begin{minipage}{0.32\textwidth}
  \centering
  \includegraphics[width=\textwidth]{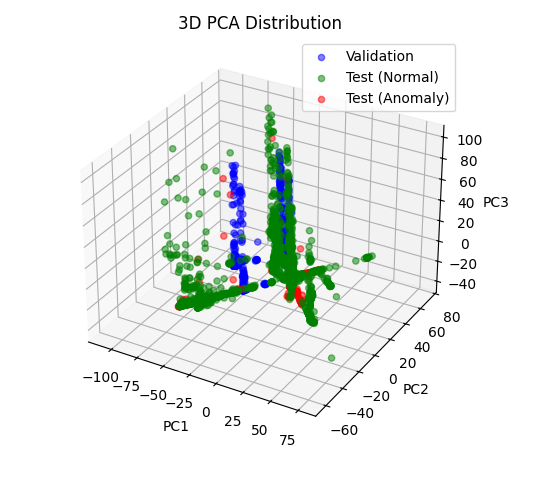} 
  \caption{Training with GSL}
  \label{fig:pca_re3}
\end{minipage}}
\end{figure}

\begin{itemize}

\item \textbf{(Original STGNN-Hyper Model):} Figure \ref{fig:pca_re1} shows a relatively scattered data distribution. There is significant overlap between normal and anomalous samples, especially in the central region of PC1. This indicates that the original model faces challenges in distinguishing certain types of anomalies. However, there are some isolated clusters of anomalies in the higher value region of PC1, indicating that the model has good recognition capabilities for specific types of anomalies.

\item \textbf{(STGNN-Hyper Model with Integrated Masking):} Figure \ref{fig:pca_re2} shows a more complex data distribution structure. The boundary between normal and anomalous samples becomes more blurred, but at the same time, anomalous samples exhibit a more scattered and diversified distribution. This may suggest that the masking mechanism enhances the model's ability to capture complex, subtle anomaly patterns, but also increases the model's complexity. More epochs of dataset traversal may be required to achieve optimal fitting results.

\item \textbf{(STGNN-Hyper Model with GSL Mechanism):} Compared to the original model, Figure \ref{fig:pca_re3} exhibits the most structured data distribution. We observe clear radial patterns, with anomalous samples tending to be distributed at the ends of these radial branches. This enables the model to better identify these anomaly points in the low-dimensional space. This structure suggests that the GSL mechanism significantly improves the model's ability to distinguish between different types of anomalies. Anomalous samples form clearer decision boundaries in the three-dimensional space.
\end{itemize}

\subsubsection{Gaussian Mixture Model}
\begin{figure}[H]
\centering
\begin{subfigure}{0.49\textwidth}
    \centering
    \includegraphics[width=\textwidth]{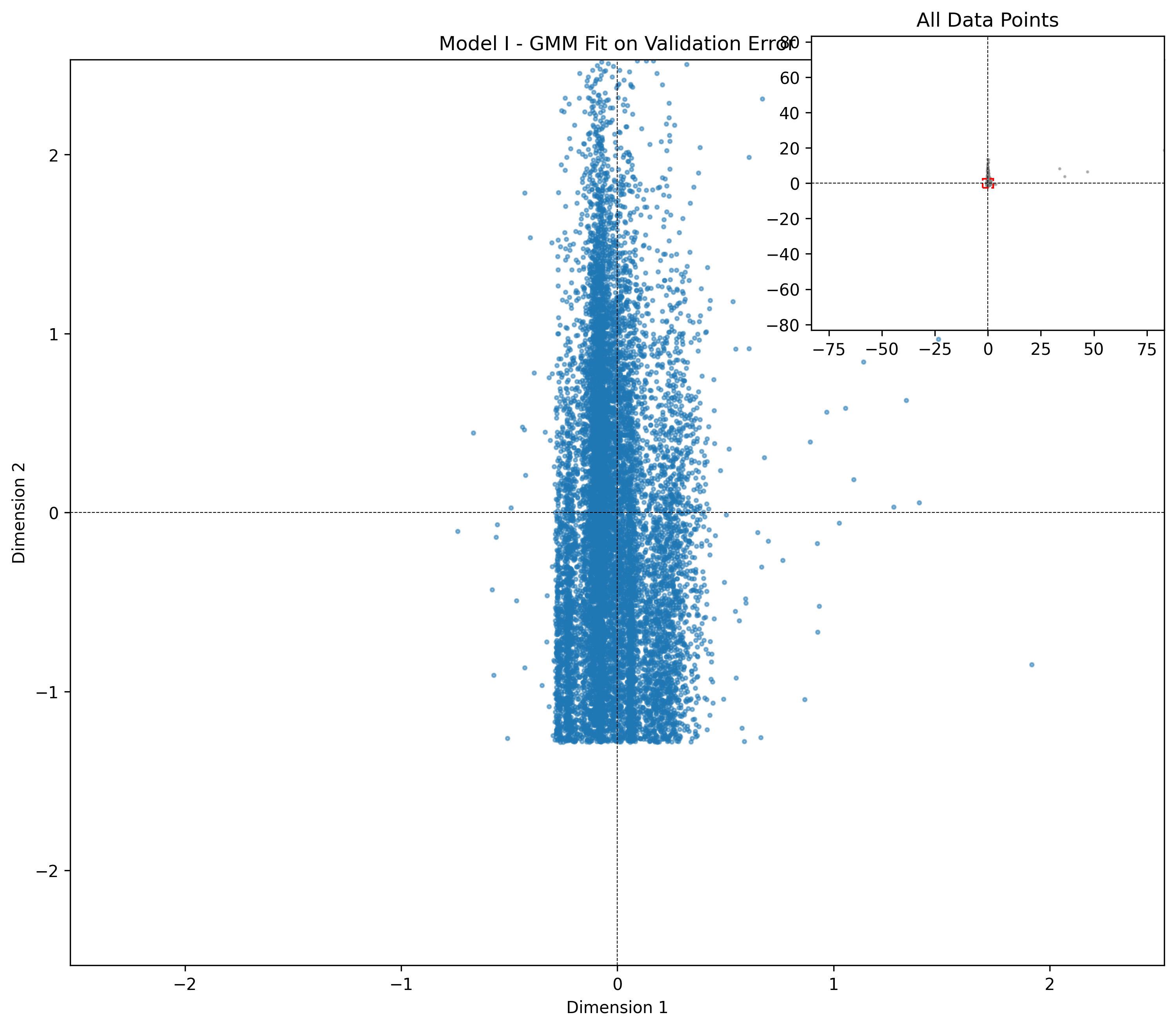} 
    \caption{Gaussian fitting on the validation set}
    \label{fig:GMM_v4}
\end{subfigure}
\begin{subfigure}{0.49\textwidth}
    \centering
    \includegraphics[width=\textwidth]{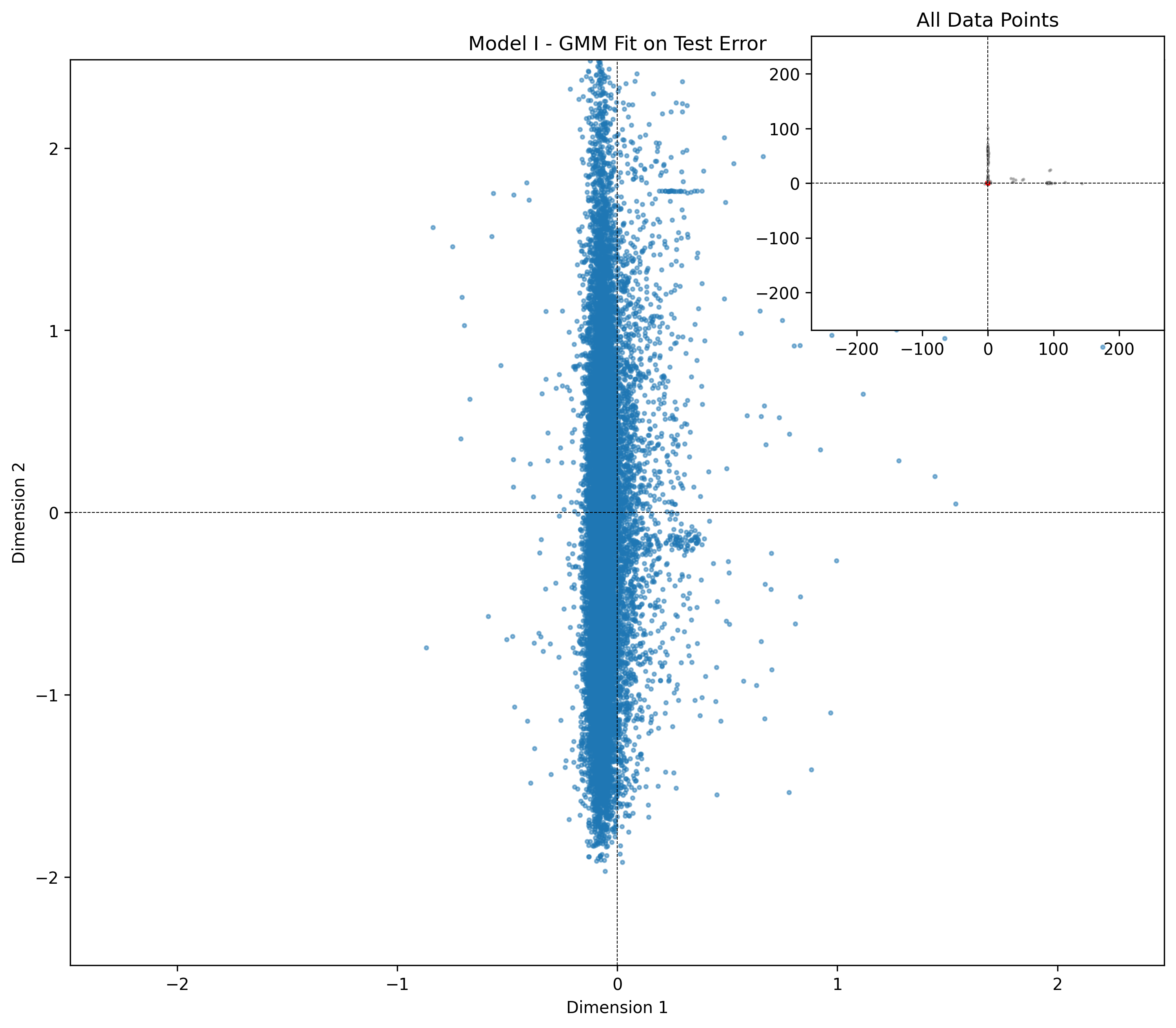} 
    \caption{Gaussian fitting on the testing set}
    \label{fig:GMM_v5}
\end{subfigure}
\vspace{10pt}  
\caption{Visualization of GMM fitting effect}
\label{fig:GMM_fitting}
\end{figure}

Figure \ref{fig:GMM_fitting} presents two subfigures showing the GMM's two-dimensional projection fitting results on the reconstruction errors of the validation set and test set, respectively. In both datasets, especially in the test set (right subfigure), we observe a significant number of outliers, which are data points far from the dense central ellipsoid regions (main components of the GMM). These points are very prominent in the visualizations. These outliers represent potential anomalies as they deviate significantly from the normal data distribution pattern captured by the GMM.

In the visualization, the red ellipses in the upper-right subfigure represent the fitted GMM components, whose sizes, shapes, and orientations reflect the local covariance structure of the data. Additionally, the Gaussian fitting effect on the test set is better, and the data points that significantly deviate from the normal pattern are clearly visible in the upper-right subfigure, mainly concentrated to the right of the normal components. This illustrates the model's generalization capability in detecting anomalies on the test data.

\section{Conclusion}
This project proposes a spatiotemporal graph convolutional neural network model based on a hypergraph structure, namely STGCN-Hyper. The model captures multivariate high-order correlations between variables through a hypergraph structure and overcomes the limitations of predefined graphs in graph learning by embedding dynamic graph structure learning. In addition, the model simultaneously captures features of different scales in the time dimension through multi-scale time dilation convolutions and supports multi-stream parallel training through CUDA Streams. This allows for the effective detection of various abnormal signal patterns in time series datasets. According to the test results of real datasets from various fields, the STGCN-Hyper model achieves anomaly detection accuracy and recall that exceed most baseline models on complex multivariate time series datasets. Therefore, the model has broad application prospects in fields such as industrial equipment monitoring, cloud server monitoring, and spacecraft status detection.
In the future, we will consider using graph attention mechanisms to enhance graph structure learning modules instead of relying on a combination of two-node embedding and linear layers. In addition, We will explore the use of more advanced models (such as autoencoder-based models) to build anomaly detectors instead of using methods such as PCA or GMM that rely on data distribution assumptions.

\appendix
\section*{Appendix}
\addcontentsline{toc}{section}{Appendix}

\section{Appendix A. Environment Dependencies and Packages Dependencies}
In this part's Appendix, we provide the operational environment, hardware facilities and the resources, dependencies used in this project that we relied on while developing the program. This information will help you reproduce the setup:

\begin{table}[H]
\centering
\begin{tabular}{|l|l|}
\hline
\textbf{Component}        & \textbf{Specification}                              \\ \hline
Operating System          & Red Hat Enterprise Linux release 8.4 (Ootpa)        \\ \hline
CPU                       & Intel(R) Xeon(R) CPU E5-2695 v4 @ 2.10GHz           \\ \hline
RAM                       & 251Gb DDR4                                          \\ \hline
GPU                       & Tesla V100-SXM2-16GB                                \\ \hline
CUDA version              & cuda11.8                                            \\ \hline
Python Version            & Python 3.9.19                                       \\ \hline
Pytorch Version           & Pytorch 2.3.1 GPU version                           \\ \hline
\end{tabular}
\vspace{10pt}  
\caption{Development Environment Specifications}
\label{tab:environment}
\end{table}

\begin{table}[H]
\centering
\begin{tabular}{|l|p{10cm}|}
\hline
\textbf{Package/Library} & \textbf{Descriptions} \\ \hline
Pytorch & 
- A deep learning framework used for building and training neural network models (especially graph neural networks). \\
& - Provides tensor operations for data manipulation. \\
& - Enables GPU acceleration for model training and inference. \\ \hline

PyTorch Geometric & 
- Provides implementations of various graph neural network layers. \\
& - Implements message passing mechanism for graph neural networks. \\ \hline

NumPy & 
- Handles array operations in data preprocessing. \\
& - Performs statistical calculations (mean, std) for data normalization. \\
& - Used in various utility functions for data manipulation. \\ \hline

Matplotlib & 
- Generates lightcurves for detected transients. \\
& - Produces animated visualizations of data slices. \\ \hline

SciPy & 
- Provides statistical functions for threshold calculations. \\ \hline

scikit-image & 
- Assists in feature detection for transient identification. \\ \hline

Astropy & 
- Reads and writes FITS files. \\
& - Handles World Coordinate System (WCS) transformations. \\
& - Performs coordinate conversions and calculations (SkyCoord). \\ \hline
\end{tabular}
\vspace{10pt}  
\caption{Packages and Libraries Dependencies}
\label{tab:packages}
\end{table}

\section{Appendix B. Anomaly Detection Performance on Low-Dimensional Datasets}
In this Appendix, we will provide more models’ performance on anomaly detection on low-dimensional time series datasets, based on the two metrics of Precision and Recall.

\begin{table}[H]
\centering
\scriptsize
\begin{tabular}{|c|c|c|c|c|c|c|c|c|}
\hline
\multirow{2}{*}{} & \multicolumn{2}{c|}{SMAP} & \multicolumn{2}{c|}{MBA} & \multicolumn{2}{c|}{SMD} \\
\cline{2-7}
 & Pre & Rec & Pre & Rec & Pre & Rec \\
\hline
LSTM-NDT & 0.8522 & 0.7325 & 0.9206 & 0.9716 & 0.7855 & 0.8528 \\

DCdetector & 0.8110 & 0.6350 & 0.xxxx & 0.xxxx & 0.6742 & 0.2189 \\

OmniAnomaly & \textbf{0.9610} & 0.7421 & 0.8560 & 0.9901 & 0.8880 & 0.9960 \\

LOF & 0.5893 & 0.5633 & 0.5789 & 0.9047 & 0.5634 & 0.3986 \\

OCSVM & 0.5385 & 0.5907 & 0.6275 & 0.8089 & 0.4434 & 0.7672 \\

MAD-GAN & 0.8156 & 0.9215 & 0.9395 & 0.9891 & 0.7290 & 0.8409 \\

DAGMM & 0.8068 & 0.6360 & 0.9474 & 0.9900 & 0.5951 & 0.8782 \\

GDN & 0.7481 & 0.9892 & 0.8833 & 0.9893 & 0.8443 & 0.4684 \\

\hline

Our model & \textbf{0.8340} & \textbf{0.9999} & \textbf{0.9654} & \textbf{0.9999} & \textbf{0.9810} & \textbf{0.9971} \\

\hline
\end{tabular}
\vspace{10pt}  
\caption{Performance of different models on low-dimensional datasets}
\label{tab:resultslow}
\end{table}

Based on the test data shown in \ref{tab:resultslow}, we can analyze the performance of the TCN-Transformer model, the TCN-Fourier model, and other baseline models on the three low-dimensional time series datasets: SMAP, MBA, and SMD.

\begin{itemize}

\item Our model exhibits the best performance across all test datasets. In the SMAP dataset, this model achieves an accuracy of 0.8340 and a recall of 0.9999. Additionally, optimized models significantly outperform other baseline models, particularly in terms of recall. However, it is noteworthy that OmniAnomaly achieves the highest accuracy (0.9610) on this dataset, but its recall (0.7421) is significantly lower than that of the TCN models.

\item On the MBA dataset, our model also performs exceptionally well, with an accuracy of 0.9654 and a recall of 0.9999. The model significantly outperform other baseline models, such as LSTM-NDT (accuracy 0.9206, recall 0.9716) and MAD-GAN (accuracy 0.9395, recall 0.9891).

\item The results on the SMD dataset are even more striking, with our model achieving an accuracy of 0.9810 and a recall of 0.9971. This performance far surpasses that of other baseline models, such as OmniAnomaly (accuracy 0.8880, recall 0.9960) and DAGMM (accuracy 0.5951, recall 0.8782).

\end{itemize}

Notably, our optimized model exhibits extremely high recall rates across all datasets while maintaining high accuracy. This indicates that the model can detect almost all abnormal signals while effectively controlling the false positive rate, achieving a good balance between precision and recall.

\bibliographystyle{plain}
\bibliography{references}

\end{document}